\documentstyle[jair,twoside,11pt,theapa,psfig]{article} \input epsf

\jairheading{9}{1998}{247-293}{5/98}{11/98}

\ShortHeadings{An Empirical Approach to Temporal Reference Resolution}
{Wiebe, O'Hara, \"{O}hrstr\"{o}m-Sandgren, \& McKeever}

\firstpageno{247}

\title{An Empirical Approach to Temporal Reference Resolution}
\author{\name Janyce M. Wiebe \email wiebe@cs.nmsu.edu \\
       \name Thomas P. O'Hara \email tomohara@cs.nmsu.edu \\
       \name Thorsten~\"{O}hrstr\"{o}m-Sandgren \email
sandgren@lucent.com \\
      \name Kenneth J. McKeever \email  kmckeeve@redwood.dn.hac.com
\\
       \addr Department of Computer Science and the Computing Research
Laboratory \\
New Mexico State University \\ Las Cruces, NM 88003}

\begin{document}
\maketitle
\begin{abstract}
Scheduling dialogs, during which people negotiate the times of
appointments, are common in everyday life.  This paper reports the
results of an in-depth empirical investigation of resolving explicit
temporal references in scheduling dialogs.  There are four
phases of this work: data annotation and evaluation, model development,
system implementation and evaluation, and model evaluation and
analysis.  The system and model were developed primarily on one set of
data, and then applied later to a much more complex data set, to
assess the generalizability of the model for the task being performed.
Many different types of empirical methods are applied
to pinpoint the strengths and weaknesses of the approach.
Detailed
annotation instructions were developed and an intercoder reliability
study was performed, showing that naive annotators can reliably perform
the targeted annotations.  A fully automatic system has been developed
and evaluated on unseen test data, with good results on both data
sets.  We adopt a pure realization of a recency-based focus model to
identify precisely when it is and is not adequate for the task being
addressed.  In addition to system results, an in-depth evaluation of
the model itself is presented, based on detailed manual annotations.
The results are that few errors occur specifically due to the model of
focus being used, and the set of anaphoric relations defined in
the model are low in ambiguity for both data sets.  

\end{abstract}
\section{Introduction}
\label{intro}
Temporal information is often a significant part of the meaning
communicated in dialogs and texts, but is often left implicit, to be
recovered by the listener or reader from the surrounding context.
When scheduling a meeting, for example, a speaker may ask, ``How about
2?'' expecting the listener to determine which day is being
specified.  Recovering temporal information implicitly communicated in
the discourse is important for many natural language processing
applications.  For example, consider extracting information from memos
and reports for entry into a database.  It would be desirable to enter
completely resolved dates and times, rather than
incomplete components such as the day or time alone.
A specific application for which temporal reference resolution is
important is appointment scheduling in natural language between 
human and machine agents \cite{busemann-etal97}.
To fully participate, the machine agent
must be able to understand the many references to times
that occur in scheduling dialogs.

Maintaining the temporal context can aid in other
aspects of understanding.
For example, Levin et al.~\citeyear{levin-etal95} and Ros\'{e} 
et al.~\citeyear{rose-etal95}
found that the temporal context, as part of the larger discourse
context, can be exploited to improve various kinds of disambiguation,
including speech act ambiguity, type of sentence ambiguity, and type
of event ambiguity.

This paper presents the results of an in-depth
empirical investigation of
temporal reference resolution.
Temporal reference resolution involves
identifying temporal information that is missing due to anaphora, and
resolving deictic expressions, which must be interpreted with respect
to the current date.  
The genre addressed is scheduling dialogs, 
in which participants schedule meetings with one another. 
Such strongly task-oriented dialogs would arise in many useful
applications, such as automated information providers and 
phone operators.

A model of temporal reference resolution in scheduling dialogs was
developed through an analysis of a corpus of scheduling dialogs.
A critical component of any method for anaphora resolution
is the focus model used.  It appeared from our initial
observations that a
recency-based model might be adequate.  
To test this hypothesis, we made the strategic 
decision to limit ourselves to a
local, recency-based model of focus, and to analyze the adequacy of 
such a model for temporal reference resolution in this
genre.  We also limit the complexity of our algorithm
in other ways.
For example, there are no facilities
for centering within a discourse segment 
\cite{sidner79,grosz-etal95}, and only very limited ones for performing tense and aspect
interpretation.  Even so, the
methods investigated in this work go a long way toward solving the
problem.  

From a practical point of view, the method is reproducible and
relatively straightforward to implement.
System results and the
detailed algorithm are presented in this paper.  
The model and the implemented system were developed primarily on one
data set, and then applied
later to a much more complex data set 
to assess the generalizability of the model for the task
being performed.  Both data sets are challenging, in that they both
include negotiation, contain many disfluencies, and show a
great deal of variation in how dates and times are discussed.
However, only in the more complex data set
do the participants discuss their real life commitments or
stray significantly from the scheduling
task.

To support the computational work,  
the temporal references in the corpus were manually
annotated.   We developed explicit annotation instructions and
performed an intercoder reliability study involving naive subjects, with
excellent results.  To support analysis of the problem and
our approach, additional manual annotations were performed, including
anaphoric chain annotations.

The system's performance on unseen test data from both data sets is
evaluated.  On both, the system achieves a large improvement over the
baseline accuracy.  In addition, ablation (degradation) experiments
were performed, to identify the most significant aspects of the
algorithm.  The system is also evaluated on unambiguous input, to help
isolate the contribution of the model itself to overall performance.

The system is an important aspect
of this work, but does not enable direct evaluation
of the model, due to errors committed by the system in
other areas of processing. 
Thus, we evaluate the model itself based on detailed
manual annotations of the data.
Important questions addressed are how many errors are
attributable specifically to the model of focus and what kinds of
errors they are, and how good is the coverage of the set of anaphoric
relations defined in the model and how much ambiguity do the
relations
introduce.  The analysis shows that few errors occur specifically
due to the model of focus, and the relations are low in
ambiguity for the data sets.

The remainder of this paper is organized as follows.
The data sets are described in Section \ref{corpora}.
The problem is defined and the results of an intercoder reliability study 
are presented in Section
\ref{intercoder}. 
An abstract 
model of
temporal reference resolution is presented in Section
\ref{model} and the high-level algorithm is presented
in Section \ref{algorithm}.
Detailed results of the implemented system are included
in Section \ref{results}, and other approaches
to temporal reference resolution are discussed 
in Section \ref{otherwork}.
In the final part of the paper, we
analyze the 
challenges presented by the dialogs to an algorithm
that does not include a model of global focus (in Section
\ref{focusmodelanal}), evaluate the coverage,
ambiguity, and correctness of the set of anaphoric relations defined
in the model (in Section \ref{relanal}), and assess the importance of
the architectural components of the algorithm
(in Section
\ref{archeval}).  Section \ref{conclusions} is the conclusion.

There are three online appendices. Online Appendix 1 contains a
detailed specification of the temporal reference resolution rules that
form the basis of the algorithm. 
Online Appendix 2 gives a specification of the input to the algorithm.
Online
Appendix 3 contains a BNF grammar describing the core set of the
temporal expressions handled by the system.
In addition, the annotation instructions, sample dialogs, and
manual annotations of the dialogs are available on the project
web site, http://www.cs.nmsu.edu/\~{  }wiebe/projects.  

\section{The Corpora} 
\label{corpora}

The algorithm was primarily developed on a sample of a corpus
of Spanish dialogs collected under the JANUS project 
at Carnegie Mellon
University \cite{shum-etal94}.
These dialogs are referred to here
as the ``CMU dialogs.''  The algorithm 
was later tested on a corpus of Spanish dialogs
collected under the Artwork project 
at New Mexico State University
by Daniel Villa and his students
\cite{wiebe-etal96}.
These are referred to here as the ``NMSU dialogs.''
In both cases, subjects were asked to set up a meeting
based on schedules given to them detailing their commitments.  
The NMSU dialogs are face-to-face, while the CMU dialogs
are like telephone conversations.
The participants in the CMU dialogs rarely discuss anything
from their real lives, and
almost exclusively
stay on task.
The participants in the NMSU dialogs
embellish the schedule given to them
with some of their real life commitments,
and often stray from the task,
discussing topics other than the
meeting being planned.

\section{The Temporal Annotations and Intercoder Reliability Study}
\label{intercoder}
Consider the passage shown in Figure \ref{corpusexample}, which is
from the CMU corpus (translated into English).
An example of temporal reference resolution is that
utterance (2) refers to 2-4pm, Thursday 30 September.

\begin{figure}
\begin{center}
\begin{tabular}{|lll|}
\hline
\multicolumn{3}{|c|}{{\it Temporal context:
Tuesday 28 September}} \\
\hline
s1& 1&   On Thursday I can only meet after two pm \\
   &2&  From two to four \\
   &3&     Or two thirty to four thirty \\
   &4&    Or three to five \\
s2& 5&   Then how does from two thirty to four thirty seem to you \\
   &6&   On Thursday \\
s1& 7&  Thursday the thirtieth of September \\
\hline
\end{tabular}
\end{center}
\caption{Corpus Example}
\label{corpusexample}
\end{figure}

Because the dialogs are centrally concerned with negotiating
an interval of time in which to hold a meeting, our
representations are geared toward such intervals.
The basic representational unit is given in Figure \ref{tempunit}. 
It is referred to throughout 
as a {\it Temporal Unit} ({\it TU}).
\begin{figure*}
\begin{center}
\vspace*{2mm}
\begin{tabular}{|lllll|}
\hline
((start-month, & start-date, & start-day-of-week, & start-hour\&minute,
& start-time-of-day) \\
\hspace*{1mm}(end-month, & end-date, & end-day-of-week, & end-hour\&minute, &
end-time-of-day))  \\
\hline
\end{tabular}
\end{center}
\caption{The Temporal Unit Representation}
\label{tempunit}
\end{figure*}

\noindent
For example, the time specified\footnote{Many terms have been used
in the literature for the relation between anaphoric expressions and
discourse
entities.  For example, 
Sidner \citeyear{sidner83} and Webber \citeyear{webber83} 
argue that ``refer'' should
be reserved for something
people do with words, rather than something words do.
Webber uses the term ``evoke'' for first references to an entity and 
``access'' for subsequent references.
Sidner uses the term ``specify'' for the relation between a noun
phrase and a discourse entity. We primarily use Sidner's term, but
use ``refer'' in a few contexts in which it seems more natural.}
in ``From 2 to 4, on Wednesday the 19th of August'' is represented as 
follows:

\begin{center}
\vspace*{2mm}
\begin{tabular}{|lllll|}
\hline
((August, & 19, & Wednesday, & 2,
& pm) \\
\hspace*{1mm}(August, & 19, & Wednesday, & 4, &
pm))  \\
\hline
\end{tabular}
\end{center}

\noindent
Thus, the information from multiple noun phrases is often
merged into a single representation of the underlying interval
specified by the utterance.  

Temporal references to times in utterances such as 
``The meeting starts at 2'' are also represented in terms of intervals.
An issue this kind of
utterance raises is whether or not a speculated end time of the
interval should be filled in, using knowledge of how long meetings
usually last.  In the CMU data, the meetings all last two hours, by design.
However, 
our annotation
instructions are conservative with respect to filling in an end
time given a starting time (or vice versa), 
specifying that it
should be left open unless something in the dialog explicitly suggests
otherwise.  This policy makes the instructions
applicable to a wider class of dialogs.

Weeks, months, and years are represented as intervals starting with
the
first day of the interval (for example, the first day of the week),
and ending with the last day of the interval (for example, the last
day of the week).

Some times
are treated as points in time
(for example, the time specified in ``It is now 3pm'').
These are 
represented as Temporal Units with the same starting and end times 
\cite<as in>{allen84}.  
If just the starting or end time is specified, all the fields of the
other end of the interval are null.
And, of course, all fields are null for
utterances that do not contain any temporal information.
In the case of an utterance that specifies multiple, distinct
intervals, the representation is a list of Temporal Units
\cite<for further details of the coding scheme, see>{ohara-etal97}.

Temporal Units are also the representations used in the evaluation 
of the system.
That is, the  system's answers
are mapped from its more complex internal representation (an
{\it ILT}, see Section \ref{architecture}) into this
simpler vector representation before evaluation is performed.

The evaluation Temporal Units used to assess the system's
performance were annotated by personnel working on the project.
The training data were annotated by the second author of this paper,
who also worked on developing the rules and other knowledge
used in the system.  However, the test
data were annotated by another project member, Karen Payne, 
who contributed
to the annotation instructions and to the integration of the
system with the Enthusiast system (see
below in Section \ref{architecture}), but did not contribute
to developing the rules and other knowledge used in the system.

As in much recent empirical work in discourse processing 
\cite<see, for example,>{arhenberg-etal95,isardcarletta95,litmanpassonneau95,mosermoore95,hirschbergnakatani96}, we performed
an intercoder reliability study investigating agreement in annotating
the times.  
The main goal in developing annotation instructions is to make them
precise but intuitive so that they can be used reliably by non-experts
after a reasonable amount of training
\cite<see>{passonneaulitman93,condoncech95,hirschbergnakatani96}.
Reliability is measured in terms of the amount of agreement among
annotators; high reliability indicates that the encoding scheme is
reproducible given multiple annotators.  In addition, the instructions
also serve to document the annotations.

The subjects were three people with no previous 
involvement in the project. They were given the original
Spanish and the English translations. However, as they have limited knowledge
of Spanish, in essence they annotated the English translations.

The subjects annotated two training dialogs according to the instructions.
After receiving feedback, they annotated four unseen test dialogs.  
Intercoder reliability was assessed using Cohen's Kappa
statistic ($\kappa$) 
\cite{siegelcastellan88,carletta96}.  Agreement for each Temporal Unit field
(for example, {\it start-month}) was assessed independently.

$\kappa$ is calculated as follows:

\[ \kappa = \frac{Pa - Pe}{1 - Pe} \]

\noindent
The numerator is the average
percentage agreement among the annotators (Pa) 
less a term for expected chance agreement (Pe), and the 
denominator is 
100\% agreement less the same term for chance agreement (Pe).

Pa and Pe are calculated as follows \cite{siegelcastellan88}.
Suppose that there are $N$ objects, $M$ classes, and $K$ taggers.
We have the following definitions.
\begin{itemize}
\item
$n_{ij}$ is  the number of assignments of object $i$ to category $j$.
Thus, for each $i$, $\sum_{j=1}^{M}{n_{ij}} = K$.   
\item
$C_j = \sum_{i=1}^{N} n_{ij}$, 
the total number of assignments of objects to category $j$. \\
\item
$p_j = \frac{C_j}{N \times K}$, the percentage of assignments
to category $j$ (note that $N \times K$ is the total number
of assignments).
\end{itemize}
We can now define $Pe$:

\[ Pe = \sum_{j=1}^{M} p^{2}_{j} \]

The extent of agreement among the taggers concerning the $ith$ object
is $S_i$, defined as follows.  It is the total number of actual agreements for object $i$,
over the maximum possible agreement for one object: 

\[ S_i = \frac{\sum_{j=1}^{M} \left( \begin{array}{c} n_{ij} \\ 2  
\end{array} \right)}
{\left( \begin{array}{c} K \\ 2  \end{array} \right)} \]

\noindent
Finally, $Pa$ is the average agreement over objects: 

\[ Pa = \frac{1}{N} \sum_{i=1}^{N} S_i \]

\noindent
$\kappa$ is 0.0 when the agreement
is what one would expect under independence, 
and it is 1.0 when the agreement
is exact \cite{hays88}.
A $\kappa$ value of 0.8 or greater indicates a high level of reliability
among raters, with values between 0.67 and 0.8 indicating only moderate
agreement \cite{hirschbergnakatani96,carletta96}.

In addition to measuring intercoder reliability, we compared each
coder's annotations to the gold standard annotations
used to assess the system's
performance.  
Results for both types of agreement are shown in Table
\ref{kappa-stats}.
The agreement among coders is shown in 
the column labeled $\kappa$, and the
average pairwise $\kappa$ values for the coders and
the expert who performed the gold standard annotations are
shown in the column labeled $\kappa_{avg}$.  
This was calculated by averaging the
individual $\kappa$ scores (which are not shown).

\begin{table*}
\begin{center}
\begin{tabular}{|l|l|l|c||c|}
\hline
Field        &      Pa &   Pe & $\kappa$ & $\kappa_{avg}$ \\
\hline					   
start        &         &      &          &     \\
Month        &     .96 &  .51 &  .93     & .94 \\
Date         &     .95 &  .50 &  .91     & .93 \\
DayofWeek      &     .96 &  .52 &  .91     & .92 \\
HourMin      &     .98 &  .82 &  .89     & .92 \\
TimeDay      &     .97 &  .74 &  .87     & .74 \\
\hline					   
end          &         &      &          &     \\
Month        &     .97 &  .51 &  .93     & .94 \\
Date         &     .96 &  .50 &  .92     & .94 \\
DayofWeek      &     .96 &  .52 &  .92     & .92 \\
HourMin      &     .99 &  .89 &  .90     & .88 \\
TimeDay      &     .95 &  .85 &  .65     & .52 \\
\hline
\end{tabular}
\caption{Agreement Among Coders}
\label{kappa-stats}
\end{center}
\end{table*}
There is a high level of agreement among
annotators in all cases except the {\it end time of day} field, 
a weakness we are
investigating.  There is also good agreement
between the evaluation annotations and the naive coders'
evaluations: with the exception of
the {\it time of day} fields, 
$\kappa_{avg}$ indicates high average pairwise agreement between the 
expert and the naive subjects.

Busemann et al.~\citeyear{busemann-etal97} 
also annotate temporal information in a corpus
of scheduling dialogs.  However, their annotations are at the level of
individual expressions rather than at the level of Temporal Units, and
they do not present the results of an intercoder reliability study.

\section{Model}
\label{model}

This section presents our model of temporal reference resolution
in scheduling
dialogs.  
Section \ref{deictic} describes the cases of deictic reference
covered and Section \ref{anaphoric} 
presents the anaphoric relations defined.
Section \ref{focusmodel} gives some background 
information
about focus
models, and then describes the focus model used in this work.

Anaphora is treated in this paper as a relationship
between a Temporal Unit representing a time specified in the current
utterance ($TU_{current}$) and one representing a time specified in a previous
utterance ($TU_{previous}$).
The resolution of the anaphor is a new Temporal Unit representing
the interpretation, in context, of the contributing words in the
current utterance.

\begin{figure*}
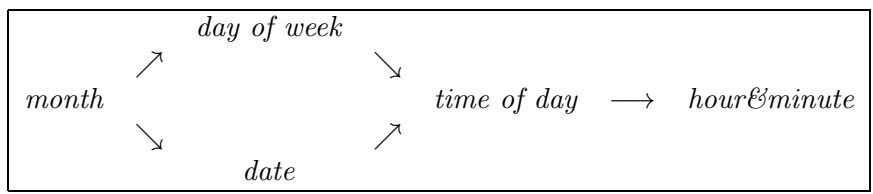

\begin{center}
\def\ne{$\nearrow$}
\def\se{$\searrow$}
\def\e{$\longrightarrow$}
\noindent
{\em
\begin{tabular}{|lrclccc|}
\hline
      &     & day of week &      &             &    &               \\
      & \ne &         & \se  &             &    &               \\
month &     &         &      & time of day & \e & hour\&minute  \\
      & \se &         & \ne  &             &    &               \\
      &     & date    &      &             &    &               \\
\hline
\end{tabular}
}
\end{center}
\caption{Specificity Ordering}
\label{specificity}
\end{figure*}

Fields of Temporal Units are partially ordered as in Figure \ref{specificity},
from least to most specific. The month has the lowest specificity
value.

In all cases of deictic reference listed in
Section \ref{deictic} and all cases of anaphoric reference
listed in Section \ref{anaphoric}, 
after the resolvent has been formed, it
is subjected to highly accurate, obvious inference to
produce the final interpretation.  Examples are
filling in the day of the week given the month and the date;
filling in {\it pm} for modifiers such as ``afternoon''; and
filling in the duration of an interval from the starting
and end points. 

In developing the rules, 
we found domain knowledge and task-specific
linguistic conventions to be most useful. 
However, we observed some cases in the NMSU data
for which syntactic information could be exploited
\cite{grosz-etal95,sidner79}.  
For example, ``until'' in the following 
suggests that the first utterance specifies an end time.
\begin{quotation}
\noindent
``... could it be until around twelve?'' \\
``12:30 there''
\end{quotation}
A preference for parallel syntactic roles
might be used to recognize that the second utterance specifies an
end time too.  We intend to pursue such preferences in future work.
\subsection{Deictic References}
\label{deictic}
The deictic expressions addressed in this work are those
interpreted
with respect to the dialog date (i.e., ``today'' in the
context of the dialog).

\subsubsection{Simple deictic relation}
A deictic expression such as ``tomorrow'' or ``last week''
is interpreted with 
respect to the dialog date.
(See rule D-simple in Section \ref{rules}.) \\

\subsubsection{Frame of reference deictic relation}
A forward time reference is calculated using the dialog date as a frame of
reference. 
Let $F$ be the most specific field in $TU_{current}$
less specific than {\it time of day} 
(e.g., the {\it date} field).
The resolvent is the next $F$ after the dialog date, augmented with
the fillers of the fields in $TU_{current}$ that are at least as
specific as
{\it time of day}.
(See rule D-frame-of-reference in Section \ref{rules}.)

Following is an example.  Assume that the dialog date is
{\it Monday 19 August}. 

\begin{center}
\begin{tabular}{|lr|}
\hline
\multicolumn{1}{|c}{Utterance} & \multicolumn{1}{c|}{Interpretation} \\
\hline
How about Wednesday at 2? & {\it 2 pm, Wednesday
21 August}\\ 
\hline
\end{tabular}
\end{center}

For both this and the {\it frame of reference} anaphoric
relation,
there are subcases for whether the starting and/or end
times are involved.  

\subsection{Anaphoric Relations}
\label{anaphoric}
Generally speaking,
many different kinds of relationships can be established between
an anaphor and its antecedent.
Examples are co-reference (``John saw Mary.  He$\ldots$''),
part-whole (``John bought a car.  The engine$\ldots$''), 
and individual-class (``John bought a truck.  They
are good for hauling$\ldots$'') \cite<see, for example,>{webber83}.
The latter two involve {\it bridging
descriptions} \cite<see, for example,>{clark77,heim82,poesio-etal97}:
some reasoning is required to infer the correct
interpretation.  
This section presents a set of anaphoric relations
that have good coverage for temporal expressions in 
scheduling dialogs (see Section
\ref{relanal} for an evaluation).
Many
temporal references involve bridging inferences, in 
the sense that
times are calculated by using the antecedent as a frame of reference
or by modifying a previous temporal interpretation.

\subsubsection{Co-reference anaphoric relation} 
The same times are specified, or $TU_{current}$ is more specific
than $TU_{previous}$. 
The resolvent contains
the union of the information in the two Temporal Units. 
(See rule A-co-reference in Section \ref{rules}.)

For example (see also (1)-(2) of the corpus 
example in Figure \ref{corpusexample}):

\begin{center}
\begin{tabular}{|lr|}
\hline
\multicolumn{1}{|c}{Utterance} & \multicolumn{1}{c|}{Interpretation} \\
\hline
How is Tuesday, January 30th?&  \\
How about 2? & {\it 2pm, Tuesday 30 January} \\
\hline
\end{tabular}
\end{center}

\subsubsection{Less-specific anaphoric relation}
$TU_{current}$ includes $TU_{previous}$, and
$TU_{current}$ is less specific than $TU_{previous}$.
Let $F$ be the most specific field in $TU_{current}$. 
The resolvent contains 
all of the information in $TU_{previous}$ of the same or lower
specificity than $F$. 
(See rule A-less-specific in Section \ref{rules}.) 

For example (see also (5)-(6) of the corpus example in 
Figure \ref{corpusexample}): 

\begin{center}
\begin{tabular}{|lr|}
\hline
\multicolumn{1}{|c}{Utterance} & \multicolumn{1}{c|}{Interpretation} \\
\hline
How about Monday at 2? & {\it Assume: 2pm, Monday 19 August}  \\
Ok, well, Monday sounds good. & {\it Monday 19
August} \\
\hline
\end{tabular}
\end{center}

\subsubsection{Frame of reference anaphoric relation}
This is the same as the {\it frame of reference} deictic relation above,
but the new time is calculated with respect to $TU_{previous}$
instead of the dialog date.   
(See rule A-frame-of-reference in Section \ref{rules}.)

Following are two examples: 

\begin{center}
\begin{tabular}{|lr|}
\hline
\multicolumn{1}{|c}{Utterance} & \multicolumn{1}{c|}{Interpretation} \\
\hline
Would you like to meet Wednesday, August 2nd? & \\
No, how about Friday at 2. & {\it 2pm, Friday 4
August}\\
\hline
\multicolumn{2}{c}{} \\
\multicolumn{2}{c}{} \\
\hline
\multicolumn{1}{|c}{Utterance} & \multicolumn{1}{c|}{Interpretation}
\\
\hline
How about the 3rd week of August? & \\
Let's see, Tuesday sounds good. & {\it Tuesday
of the 3rd week in August} \\  
\hline
\end{tabular}
\end{center}

In the first example, the day specified in the first utterance is
used
as the frame of reference.
In the second example, the beginning day of the interval representing
the 3rd week of August is used as the frame of reference. 

Note that tense can influence the choice of whether to calculate a
forward or backward time from a frame of reference
\cite{kampreyle93}, 
but this is not accounted for because there is not much
tense variation in the CMU corpus on which the algorithm was
developed.  However, errors can occur because backward calculations
are not covered. For
example, one might mention ``Friday'' and then ``Thursday,'' intending
``Thursday'' to be calculated as the day {\bf before} that Friday,
rather than the Thursday of the week following that Friday.
We are investigating creating a new anaphoric relation to cover these
cases. 

\subsubsection{Modify anaphoric relation}
$TU_{current}$ is calculated by modifying the 
interpretation
of the previous temporal reference.
The times
differ in the filler of a field $F$, where $F$ is at least as specific as 
{\it time of day}, but are
consistent in all fields
less
specific than $F$. 
The resolvent contains the 
information in $TU_{previous}$ that is less specific than $F$
together with the information in $TU_{current}$ that is of the same or
greater specificity as $F$.
(See rule A-modify in Section \ref{rules}.)

For example (see also (3)-(5) of the corpus example 
in Figure \ref{corpusexample}): 

\begin{center}
\begin{tabular}{|lr|}
\hline
\multicolumn{1}{|c}{Utterance} & \multicolumn{1}{c|}{Interpretation} \\
\hline
Monday looks good. & {\it Assume: Monday 19 August}  \\
How about 2? & {\it (co-reference relation) 2pm, 
Monday 19 August} \\
Hmm, how about 4? & {\it (modify relation) 4pm, Monday
19 August} \\
\hline
\end{tabular}
\end{center}

\subsection{Focus Models}
\label{focusmodel}

The focus model, or model of attentional state \cite{groszsidner86},
is a model of which entities the dialog is most centrally about at
each point in the dialog.  It determines which previously mentioned
entities are the candidate antecedents of anaphoric references.  As
such, it represents the role that the structure of the discourse
plays in reference resolution.

We consider three models of attentional state in this paper: (1)
the linear-recency model (see, for example, the work by Hobbs
\citeyear{hobbs78} and Walker\footnote{Note that Walker's model
is a cache-based model for which recency is
a very important but not unique criterion for determining
which entities are in the cache.} \citeyear{walker96}),
(2) Grosz and Sidner's \citeyear{groszsidner86} stack-based
model, and (3) the {\it graph structured stack} model introduced by
Ros\'e, Di Eugenio, Levin, and Van Ess-Dykema
\citeyear{rose-etal95}.  Ordered from (1) to (3), the models are
successively more complex, accounting for increasingly more complex
structures in the discourse.  

\begin{figure}
\begin{center}
\centerline{
\psfig{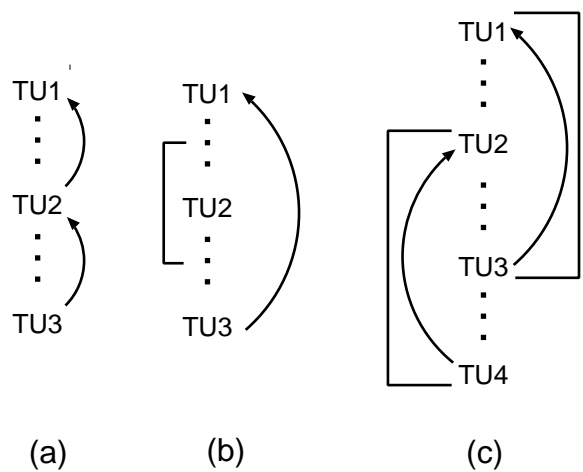}}
\caption{Discourse Structures Targeted by Different Focus Models}
\label{structures}
\end{center}
\end{figure}

In a linear-recency based model,
entities mentioned in the discourse are stored on a focus list,
ordered by recency.  The corresponding structure in the dialog is
shown in Figure \ref{structures}a: a simple progression of
references, uninterrupted by subdialogs.  

In Grosz and Sidner's
stack-based model, the entities in focus in a particular discourse
segment are stored together in a {\it focus space} associated
with that segment.  
To handle anaphoric
references across discourse segments, focus spaces are pushed
on and popped off the stack as appropriate to mirror the
structure of the discourse.  As each new segment is
recognized, a focus space is created and pushed onto the stack.
To interpret an anaphoric reference, the entities in the focus
space on the top of the stack are considered first.  However, if the 
current utterance resumes a previous discourse segment, the
intervening focus spaces are popped off.  This allows anaphoric reference
to an earlier entity, even if more recently mentioned entities
are possible antecedents
\cite<for more details, see>{groszsidner86}. 
Figure \ref{structures}b illustrates
a discourse structure that the stack-based model is designed to handle.
Suppose that both $TU_1$ and $TU_2$ are possible antecedents of $TU_3$
(for example, suppose they are specified by pronouns that agree in
number
and gender), 
but
$TU_2$ is in a subsegment and is not a correct
antecedent of $TU_3$, even though it is mentioned more recently than
$TU_1$.
In the stack-based model, the focus space containing $TU_2$ is popped
off the stack when the end of its segment is recognized, 
thus removing $TU_2$ as a competitor for understanding $TU_3$.
Following is an example from the NMSU corpus
(this is the dialog segment labeled 09-09, in row 7, in
Figure \ref{summaryfigure} presented later).

\begin{center}
\begin{tabular}{|llll|}
\hline
\multicolumn{4}{|c|}{{\it Dialog Date:  Monday 10 May}}  \\
\hline
       & 1 & S1 & Listen, daughter, I was thinking of inviting you 
to a demonstration on  \\
       &   &    & interior things, ornaments for decorating your house. \\
       & 2 &    & Uh, I would like to do it at two p.m. Wednesday, \\
       & 3 &    & But I don't know if you are free at that time or $\ldots$ \\
$TU_1$ & 4 & S2 & Uh, Wednesday, Mom, well 
\\
       &   &    & {\it Resolved to Wednesday, May 12} \\
       & 5 &    &   You know that, 
\\
$TU_{2,1}$ & 6 &  & last week uh, I got a job and uh, a full-time job
  \\
       &   &    &  {\it Unambiguous deictic; resolved to the week
before the dialog date} \\
$TU_{2,2} $ &  7 &  &  I go in from seven
in the morning to five in the afternoon \\
       &   &    &  {\it Habitual} \\
     & 8 & S1 &   Oh, maybe it would be better
 \\
$TU_3$ & 9 & S2 & Well, I have lunch from twelve to one  \\
       &   &    & {\it Utterance (4) is needed for the correct
interpretation:} \\
       &   &    & {\it 12-1, Wednesday 12 May} \\
\hline
\end{tabular}
\end{center}

\noindent
In this passage, utterances (6)-(7) are 
in a subdialog about S2's job.  
To interpret ``twelve to one''
in utterance (9) correctly, one must go back to 
utterance (4).   Incorrect interpretations 
involving the temporal references in (6) and (7) are possible
(using the {\it co-reference}
relation with (6) and the {\it modify} relation with (7)), so
those utterances must be skipped. 

Ros\'e et al.'s graph structured stack is designed to handle
the more complex structure depicted in Figure \ref{structures}c.
We will return to this structure later in Section \ref{focusmodelanal},
when the adequacy of our focus model is analyzed.

Once the candidate antecedents are determined, various
criteria can be used to choose among them.  
Syntactic and semantic constraints
are common.
\subsubsection{Our Focus Model for
Temporal Reference Resolution}

As mentioned earlier, our algorithm for temporal reference 
resolution is recency based.
Specifically, the focus model is structured as a
linear list of all times mentioned so far in the current dialog.  The
list is ordered by recency, and no entries are deleted from the 
list.

The candidate antecedents are as follows. For each type of anaphoric
relation, the most recent Temporal Unit on the focus
list that satisfies that relation,
if there is one, is a candidate antecedent.

The antecedent is chosen from among the candidate antecedents
based on a combined score reflecting a priori preferences for the type
of anaphoric relation established, how recently the time was
mentioned, and how plausible the resulting temporal interpretation
would be (see Section \ref{algorithm}).  
These numerical heuristics contribute to some extent to the
success of the implementation, 
but are not critical components of the model,
as shown in Section \ref{archeval}.

\subsection{The Need for Explicit Identification of Relations}
\label{need}
As mentioned in the introduction, one goal of this work is to assess
the adequacy of a recency-based focus model for this task and genre.
To be well founded, such evaluations must be made
with respect to a particular set of relations.
For example, the {\it modify} relation 
supports a recency-based approach.
Consider the following example, reproduced from 
Section \ref{anaphoric}:

\begin{center}
\begin{tabular}{|lr|}
\hline
\multicolumn{1}{|c}{Utterance} & \multicolumn{1}{c|}{Interpretation} \\
\hline
(1) Monday looks good. & {\it Assume: Monday 19 August}  \\
(2) How about 2? & {\it (co-reference relation) 2pm, 
Monday 19 August} \\
(3) Hmm, how about 4? & {\it (modify relation) 4pm, Monday
19 August} \\
\hline
\end{tabular}
\end{center}

Because our model includes the {\it modify} anaphoric relation, the Temporal
Unit in (2) is an appropriate antecedent for the one in (3).  A model
without this relation might require 
(3)'s antecedent to be provided by
(1).  
\section{Algorithm}
\label{algorithm}
This section presents our high-level algorithm for temporal reference
resolution.  
After an overview in Section \ref{algoverview}, the rule application
architecture is described in Section \ref{architecture}, and the
main rules composing
the algorithm are given in Section \ref{rules}.  The complete set of
rules is given in detail in Online 
Appendix 1.

\subsection{Overview}
\label{algoverview}

An important feature of our approach is that the system is
forced to choose among possibilities only if the resulting
interpretations would be inconsistent.  If the results for two
possibilities are consistent, the system merges the results together.

At a high level, the algorithm operates as follows.
There is a set of rules
for each of the relations presented
in Section \ref{anaphoric}.
The rules include constraints involving the
current
utterance and another Temporal Unit.   In the anaphoric cases, the
other 
Temporal Unit
is a potential antecedent from the focus list. In the deictic cases, it is the dialog
date or a later time.  
For the 
current temporal expression to be resolved, each rule is applied.  For
the anaphoric rules, the antecedent
considered is the most recent one satisfying 
the constraints.
All consistent
maximal mergings of the results are formed, and the one with the
highest score is the chosen interpretation.

\subsection{Architecture}
\label{architecture}
\begin{figure}
\begin{center}
\centerline{
\psfig{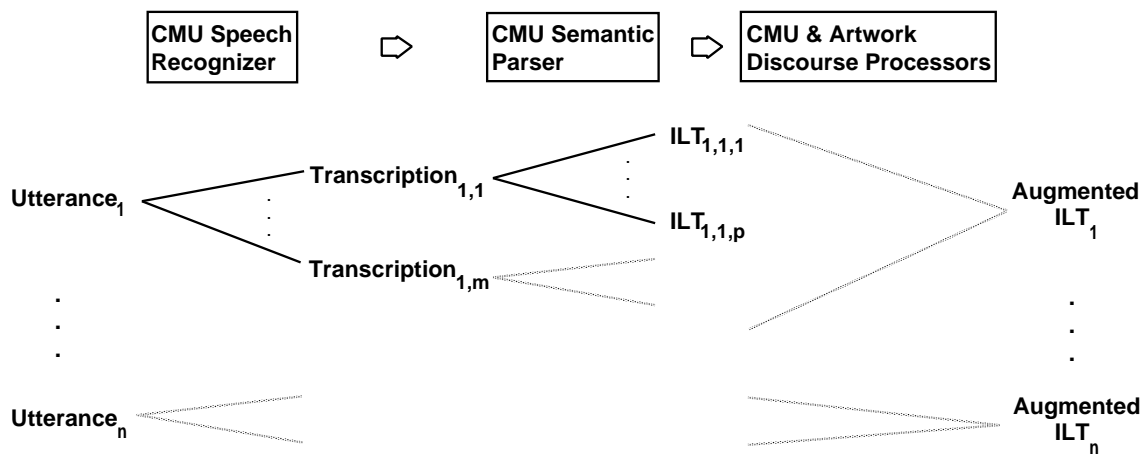}}
\caption{The Enthusiast System}
\label{enthusiast}
\end{center}
\end{figure}
Our system was developed to be integrated into the Enthusiast system
developed at Carnegie Mellon University 
\cite<see>{qu-etal96,levin-etal95,rose-etal95,lavietomita93}.  
Enthusiast is a speech-to-speech machine translation system
from Spanish into English.  The aspects of the system needed for this
paper are shown in Figure \ref{enthusiast}.
The system processes all the utterances of a single speaker turn
together (utterances 1 through $n$ in the figure). 
Each spoken Spanish utterance is input to the speech
recognizer, which produces one or more transcriptions 
of the utterance.
The output of the speech recognition system is the input to a semantic
parser \cite{lavietomita93,levin-etal95},
which produces a representation of the literal meaning of the
sentence.  This representation is called 
an {\it Interlingual Text} ({\it ILT}).
The output of the semantic parser is ambiguous, consisting of multiple
ILT representations of the input transcription.
All of the ILT representations produced for an utterance
are input to the discourse processor, which
produces the final, unambiguous representation of that utterance.
This representation is called an
{\it augmented ILT}.

The discourse processor
can be configured to be our system alone, a
plan-based discourse processor developed at CMU \cite{rose-etal95},
or the two working together in integrated mode.  
The main results, presented in Tables \ref{cmu-eval} and
\ref{nmsu-eval} in Section \ref{results}, are for our system working
alone, taking as input the ambiguous output of the semantic parser.
For the CMU dialogs, the input to the semantic parser is the output of
the speech recognition system.  The NMSU dialogs were input to the
semantic parser directly in the form of 
transcriptions.\footnote{The semantic parser but not the speech recognizer was
available for us to process the NMSU data.  
Presumably, the
speech recognizer would not perform as well on the NMSU dialogs as
it does on the CMU dialogs,
since it was trained on the latter.}

To produce one ILT, the semantic parser maps the main event and its
participants into one of a small set of case frames (for example, a
{\it meet} frame or an {\it is busy} frame).  It also produces a
surface representation of the temporal information in the utterance,
which mirrors the
form of the input utterance.  Although the events and states discussed
in the NMSU data are often outside the coverage of this parser, the
temporal information generally is not. Thus, the parser provides 
a sufficient input representation for our purposes on both sets
of data.  

As the Enthusiast system is configured, the input is presented
to our discourse processor in the form of alternative sequences of
ILTs.  Each sequence contains one ILT for each utterance.  For
example, using the notation in Figure \ref{enthusiast}, a sequence
might consist of $ILT_{1,2,3}$, $ILT_{2,1,1}$, $\ldots$, $ILT_{n,2,1}$.
Our system resolves the ambiguity in batches.  
Specifically, it
produces a sequence of Augmented ILTs for each input sequence, and
then chooses the best sequence as its final interpretation of the
corresponding utterances.  In this way, the input ambiguity is
resolved as a function of finding the best temporal interpretations of
the utterance sequences in context \cite<as suggested by>{qu-etal96}.
However, the number of alternative sequences of ILTs for a set of
utterances can be prohibitively large for our system.  The total
number of sequences considered by the system is limited to the top
125, where the sequences are ordered using statistical rankings
provided by the Enthusiast system.

Our method for performing semantic disambiguation is appropriate for
this project, because the focus is on temporal reference resolution
and not on semantic disambiguation.  However, much semantic ambiguity
cannot be resolved on the basis of the temporal discourse context
alone, so this represents a potential area for improvement in the
system performance results presented in Section \ref{results}.  
In fact, the Enthusiast researchers have already developed better
techniques for resolving the semantic ambiguity in these dialogs
\cite{shum-etal94}.

Because the ILT representation was designed to support various projects in
discourse, semantic interpretation, and machine translation, the
representation produced by 
the semantic parser is much richer than is required for our
temporal reference resolution algorithm.  We recommend that others
who implement our algorithm for their application build an input parser to
produce only the necessary temporal information. The specification of
our input is available in Online Appendix 2.

As described in Section \ref{focusmodel},
a focus list records the Temporal Units that have been discussed so far in the
dialog.  After a final Augmented ILT has been created for the current
utterance, the Augmented ILT and the utterance are placed together on the focus
list.
In the case of utterances that specify more than one
Temporal Unit, 
a separate entity is added for each to the focus list,
in order of mention.  
Otherwise, the system architecture is similar
to a standard production system, with one major exception: rather than
choosing the results of just one of the rules that fires,
multiple results can be merged.  This is a
flexible architecture that accommodates sets of rules targeting
different aspects of the interpretation.

Following are the basic steps in processing a single ILT. \\

\noindent
{\bf Step 1.}  The input ILT is {\it normalized}.  
In producing the ILTs that serve as input to our system,
the semantic parser often
represents pieces of information about the same time 
separately, mirroring the surface form of the utterance.
This is done in order to capture 
relationships, such as topic-comment relationships, among clauses.  
Our
system needs to know which pieces of information are about the same
time, but does not need to know about the additional relationships.
Thus, the system maps the input representation into a normalized
form, 
to shield the reasoning component from the idiosyncracies of
the input representation.  A specification of the normalized form
is given in Online Appendix 2.

The goal of the normalization process is to produce one Temporal
Unit per distinct time specified in the utterance.
The normalization program is quite detailed (since it
must account for the various structures possible in the
CMU input ILT), but the core strategy is straightforward:
it merges information provided by
separate noun phrases into one Temporal Unit, 
if it is consistent to do so.  Thus, new Temporal Units
are created only if necessary.  Interestingly,
few errors result from this process.
Following are some examples.
\begin{center}
\begin{tabular}{|ll|}
\hline
I can meet Wednesday or Thursday. & {\it Represented as two
disjoint TUs}. \\
I can meet from 2:00 until 4:00 on the 14th. & {\it Represented as one TU.}  \\
I can meet Thursday the 11th of August. & {\it Represented as one
TU.} \\
\hline
\end{tabular}
\end{center}

\noindent
After the normalization process, highly accurate, obvious inferences are
made and added to the representation. \\

\noindent
{\bf Step 2.}
All of the rules are applied to the normalized input.  
The result of a rule
application is a {\it Partial Augmented ILT}---information this
rule will contribute to the interpretation of the utterance, if
it is chosen.  This
information includes a certainty factor representing an a priori
preference for the type of anaphoric or deictic relation being
established. In the case of anaphoric relations, this factor is
adjusted by a term representing how far back on the focus list the
antecedent is (in the anaphoric rules in Section \ref{rules}, the adjustment
is represented by {\it distance factor} in the calculation of the
certainty factor {\it CF}).
The result of this step is the set of Partial Augmented ILTs produced
by the rules that fired (i.e., those that succeeded). 

In the case of multiple Temporal Units in the input ILT, each rule is
applied as follows.  If the rule does not access the focus list, the
rule is applied to each Temporal Unit. A list of Partial Augmented
ILTs is produced, containing one entry for each successful match,
retaining the order of the Temporal Units in the original input.  If
the rule does access the focus list, the process is the same, but with
one important difference.  The rule is applied to the
first Temporal Unit.  If it is successful, then the same focus list entity
used to apply the rule to this Temporal Unit is used to interpret the
remaining Temporal Units in the list.  Thus, all the anaphoric
temporal
references in a single utterance are understood with respect to the
same focus list element.  So, for example, the anaphoric
interpretations of the temporal expressions in
``I can meet Monday or Tuesday'' both have to be understood with
respect
to the same entity in the focus list.  

When accessing entities on the focus list, an entry for an utterance
that specifies multiple Temporal Units may be encountered.  In this
case, the Temporal Units are simply
accessed in order of mention (from most to least
recent).\\

\noindent
{\bf Step 3.}
All maximal mergings of the Partial Augmented ILTs 
are created.
Consider a graph in which the Partial Augmented ILTs are the vertices, and there is
an edge between two Partial Augmented ILTs if they are compatible.  Then,
the maximal cliques of the graph (i.e., the maximal complete subgraphs)
correspond to the maximal mergings.
Each maximal merging is then merged with the normalized
input ILT,
resulting in a set of Augmented ILTs.  \\

\noindent
{\bf Step 4.}
The Augmented ILT chosen is the one with the highest certainty factor.
The certainty factor of an Augmented ILT is calculated as follows.
First, the certainty factors of the constituent Partial Augmented ILTs
are summed. Then,
critics are applied
to the resulting Augmented ILT, 
lowering the certainty factor if the information is judged
to be incompatible with the dialog state.  

The merging process might have yielded additional
opportunities for making obvious inferences, so this process
is performed again, to produce the final Augmented ILT.

To process the alternative input sequences, 
a separate invocation
to the core system is made for each sequence, with the sequence
of ILTs and the current focus list as input.
The result of each call is a sequence of Augmented ILTs, which
are the
system's best interpretations of the input ILTs,
and a new focus list, representing the updated discourse context
corresponding to that sequence of interpretations.
The system assigns a certainty factor to each sequence of
Augmented ILTs, specifically, the sum of the certainty factors
of the constituents.  It chooses the sequence with the highest
certainty factor, and updates the focus list to the focus list
calculated for that sequence.

\subsection{Temporal Reference Resolution Rules}
\label{rules}
Figure \ref{temporal-rules} 
presents the main temporal resolution rules,
one for each of the cases described in Sections \ref{deictic} and
\ref{anaphoric}.  In the complete set of rules, given in Online Appendix 1,
many are broken down into subcases involving, for example,
the end times or starting times.

The rules apply to individual Temporal Units.
They return a certainty factor, and either a more
fully specified Temporal Unit or an empty structure indicating
failure.

Many of the rules calculate temporal information with respect to a
frame of reference, using a separate calendar utility.  
Following are functions and conventions used in Figure
\ref{temporal-rules}.

\begin{enumerate}
\item {\bf next}($TimeValue$, $RF$): returns the 
next $timeValue$ that follows reference frame $RF$. For example,
next(Monday, [$\ldots$Friday, 19th,$\ldots$]) = Monday, 22nd.

\item {\bf resolve\_deictic}($DT$, $RF$): 
resolves the 
deictic term $DT$ with respect to the reference frame $RF$.

\item {\bf merge}($TU_1$, $TU_2$):
if Temporal Units $TU_1$ and $TU_2$ contain no conflicting fields,
returns a Temporal Unit containing all of the information
in the two units; otherwise returns $\{ \}$.

\item {\bf merge\_upper}($TU_1$, $TU_2$):
similar to the previous function, except that the only fields from
$TU_1$ that
are included
are those that are of the same or less specificity 
as the most specific field in $TU_2$.

\item {\bf specificity}($TU$):
returns the specificity of the most specific field in $TU$.

\item {\bf most\_specific}($TU$): returns the most specific field in $TU$.

\item {\bf starting\_fields}($TU$): returns a list of starting field
names for those in $TU$ having non-null values.

\item {\bf structure$\rightarrow$component:}
returns the named component of the structure.

\item {\bf conventions:}
Values are in {\bf bold face} and variables are in $italics$.
$TU$ is the current Temporal Unit being resolved.
$TodaysDate$ is a representation of the dialog date. $FocusList$ is
the list of discourse entities from all previous utterances.
\end{enumerate}

{
\begin{figure}

\begin{center}
{\large \em Rules for deictic relations}
\end{center}

\begin{tabbing}
mm\=mm\=mm\=mm\=mm\=mm\=mm\=mm\= \kill
{\em \bf Rule D-simple:} All cases of the {\it simple}
deictic relation. \\
if there is a deictic term, $DT$, in $TU$ then \\
\>    return $\langle$0.9, merge($TU$,
resolve\_deictic($DT$, {\bf $TodaysDate$}))$\rangle$ \\
 \\
{\em \bf Rule D-frame-of-reference:} The starting-time cases of the
{\it frame of reference} deictic relation. \\
if (most\_specific(starting\_fields($TU$)) $<$ {\bf time\_of\_day}) then \\
\> Let $f$ be the most specific field in starting\_fields($TU$) \\
\> return 
$\langle$0.4, merge($TU$, next($TU$$\rightarrow$$f$, $TodaysDate$))$\rangle$
\end{tabbing}
\begin{center}
{\large \em Rules for anaphoric relations}
\end{center}

\begin{tabbing}
mm\=mm\=mm\=mm\=mm\=mm\=mm\=mm\= \kill
{\em \bf Rule A-co-reference:} All cases of the {\it co-reference}
anaphoric relation. \\
for each non-empty Temporal Unit $TU_{fl}$ from $FocusList$ (starting with most recent) \\
\> if specificity($TU_{fl}$) $\leq$ specificity($TU$) 
         and not empty merge($TU_{fl}$, $TU$) then \\
\>  \> $CF$ = 0.8 $-$ distance\_factor($TU_{fl}$, {\bf $FocusList$}) \\
\>  \> return $\langle$$CF$, merge($TU_{fl}$, $TU$)$\rangle$ \\
 \\

{\em \bf Rule A-less-specific:} All cases of the {\it less-specific} anaphoric 
relation. \\
for each non-empty Temporal Unit $TU_{fl}$ from $FocusList$ (starting with most recent) \\
\> if specificity($TU_{fl}$) $>$ specificity($TU$) 
      and not empty merge\_upper($TU_{fl}$, $TU$) then \\
\> \> $CF$ = 0.5 $-$ distance\_factor($TU_{fl}$, {\bf $FocusList$}) \\
\> \> return $\langle$$CF$, merge\_upper($TU_{fl}$, $TU$)$\rangle$ \\
 \\

{\em \bf Rule A-frame-of-reference:} 
Starting-time case of the {\it frame of reference} anaphoric relation. \\
if (most\_specific(starting\_fields($TU$)) $<$ {\bf time\_of\_day}) then \\
\> for each non-empty Temporal Unit $TU_{fl}$ from $FocusList$ (starting with most recent) \\
\> \> if specificity($TU$) $\geq$ specificity($TU_{fl}$) then \\
\> \> \> Let $f$ be the most specific field in starting\_fields($TU$) \\
\> \> \> $CF$ = 0.6 $-$ distance\_factor($TU_{fl}$, {\bf $FocusList$}) \\
\> \> \> return 
$\langle$$CF$, merge($TU$, next($TU$$\rightarrow$$f$, $TU_{fl}$$\rightarrow$start\_date))$\rangle$ \\
 \\

{\em \bf Rule A-modify:} All cases of the {\it modify} anaphoric relation. \\
if (specificity($TU$) $\geq$ {\bf time\_of\_day}) then \\
\> for each non-empty Temporal Unit $TU_{fl}$ from $FocusList$ (starting with most recent) \\
\> \> if specificity($TU$) $\geq$ specificity($TU_{fl}$) and
specificity($TU_{fl}$) $\geq$ {\bf time\_of\_day} then \\
\> \> \> if not empty merge\_upper($TU_{fl}$, $TU$) then \\
\> \> \> \> $CF$ = 0.5 $-$ distance\_factor($TU_{fl}$, {\bf $FocusList$}) \\
\> \> \> \> return $\langle$$CF$, merge\_upper($TU_{fl}$, $TU$)$\rangle$ \\
\end{tabbing}
\caption{Main Temporal Resolution Rules}
\label{temporal-rules}
\end{figure}
}

The algorithm does not cover some subcases of relations
concerning the end times.  
For instance, rule D-frame-of-reference covers only the
starting-time case of the {\it frame of reference} deictic relation. 
An example of an end-time case that is not handled is
the utterance ``Let's meet until Thursday,''
under the meaning that they should meet
from today through Thursday.    This is an area for future
work.
\section{Results}
\label{results}
As mentioned in Section \ref{intercoder}, the main results are based on
comparisons against human annotation of the held out test data. 
The results are based on straight
field-by-field comparisons of the Temporal Unit representations
introduced in Section \ref{intercoder}. To be considered correct,
information must not only be right, but it also has to be in the
right place.
Thus, for example, ``Monday'' correctly resolved to {\it Monday 19
August}, but incorrectly treated as a starting rather than an end
time, contributes 3 errors of omission and 3 errors of commission (and
receives no credit for the correct date).

Detailed results for the test sets are presented in this section, 
starting with
results for the CMU data (see Table \ref{cmu-eval}).  
{\it Accuracy} measures the extent to which the
system produces the correct answer, while {\it precision} measures the
extent to which the system's answers are correct (see the formulas
in Table \ref{cmu-eval}).
For each component
of the extracted temporal structure, the system's correct
and incorrect answers were counted.
Since null values occur quite often, these counts
exclude cases in which the system's answer, the correct
answer, or both answers are null.
Those cases were counted separately.
Note that
each test set contains three complete dialogs with an average of
72 utterances per dialog.

{
\begin{table*}
\noindent
\begin{center}
\begin{tabular}{|l|r|r|r|r|r||r|r|c|r|r|}
\hline 
Label     & Cor & Inc & Mis & Ext & Nul & Poss & Act & BaseAcc &  {\bf Acc}  & {\bf Prec} \\ 
\hline 
start &  &  &  &  &  &  &  &  &  & \\ 
Month     &  49 &   3 &   7 &   3 &   0 & 59 & 55 & 0.338 & 0.831 & 0.891 \\ 
Date      &  48 &   4 &   7 &   3 &   0 & 59 & 55 & 0.403 & 0.814 & 0.873 \\ 
DayofWeek &  46 &   6 &   7 &   3 &   0 & 59 & 55 & 0.242 & 0.780 & 0.836 \\ 
HourMin   &  18 &   0 &   7 &   0 &  37 & 62 & 55 & 0.859 & 0.887 & 1.000 \\ 
TimeDay   &   9 &   0 &  18 &   0 &  35 & 62 & 44 & 0.615 & 0.710 & 1.000 \\ 
\hline 
end &  &  &  &  &  &  &  &  &  & \\ 
Month     &  48 &   3 &   7 &   1 &   3 & 61 & 55 & 0.077 & 0.836 & 0.927 \\ 
Date      &  47 &   5 &   6 &   3 &   1 & 59 & 56 & 0.048 & 0.814 & 0.857 \\ 
DayofWeek &  45 &   7 &   6 &   3 &   1 & 59 & 56 & 0.077 & 0.780 & 0.821 \\ 
HourMin   &   9 &   0 &   9 &   0 &  44 & 62 & 53 & 0.862 & 0.855 & 1.000 \\ 
TimeDay   &   4 &   0 &  13 &   1 &  44 & 61 & 49 & 0.738 & 0.787 & 0.980 \\ 
\hline 
{\bf Overall} & 323 &  28 &  87 &  17 & 165 & 534 & 604 & 0.428 & {\bf 0.809} &
{\bf 0.916} \\ 
\hline 
\end{tabular}
\end{center}

\begin{center}
\begin{tabular}{ll}
{\bf Legend}   & \\
Cor(rect):     & System and key agree on non-null value \\
Inc(orrect):   & System and key differ on non-null value \\
Mis(sing):     & System has null value for non-null key \\
Ext(ra):    & System has non-null value for null key \\
Nul(l):        & Both system and key give null answer \\
\\
Poss(ible): & Correct + Incorrect + Missing + Null \\
Act(ual): & Correct + Incorrect + Extra + Null \\
Base(line)Acc(uracy): & Baseline accuracy (input used as is) \\
Acc(uracy): & \% Key values matched correctly
 ((Correct + Null)/Possible) \\
Prec(ision):  & \% System answers matching the key
 ((Correct + Null)/Actual) \\
\end{tabular}
\caption{Evaluation of System on CMU Test Data}
\label{cmu-eval}
\end{center}
\end{table*}
}

These results show that the system achieves an overall accuracy
of 81\%, which is significantly better than the baseline accuracy (defined
below) of 43\%.
In addition, the results show a high precision of 92\%.  In
some of the individual cases, however, the results could be higher due
to several factors.  For example, our system development was
inevitably focused more on some fields than others.  
An obvious area for improvement is the system's 
processing of the {\it time of day} fields.
Also, note
that the values in the {\it Mis} column are higher than those in the
{\it Ext} column.  This reflects the conservative coding convention,
mentioned in Section \ref{intercoder},  
for filling in unspecified end points.

The accuracy and precision figures for the {\it hour \& minute} and {\it
time of day} fields are very high because a large proportion of them are
null.  We include null correct answers in our figures because 
such answers often reflect valid decisions not to fill in
explicit values from previous Temporal Units.

Table \ref{nmsu-eval} contains the results for the system on the NMSU
data. It shows that the system performs respectably, with 69\%
accuracy and 88\% precision, on the more complex set of data. The
precision is still comparable, but the accuracy is lower, since more of
the entries are left unspecified (that is, the figures in the
{\it Mis} column in Table \ref{nmsu-eval} are higher than in
Table \ref{cmu-eval}).
Furthermore, the baseline
accuracy (29\%) is almost 15\% lower than the one for the CMU data
(43\%), supporting the claim that this data set is more challenging.

{
\begin{table*}
\begin{center}
\noindent
\begin{tabular}{|l|r|r|r|r|r||r|r|c|r|r|}
\hline 
Label     & Cor & Inc & Mis & Ext & Nul & Poss & Act & BaseAcc &  {\bf Acc}  & {\bf Prec} \\ 
\hline 
start &  &  &  &  &  &  &  &  &  & \\ 
TimeDay   &  9 & 0 & 18 & 0 & 35 & 62 & 44 & 0.615 & 0.710 & 1.000 \\ 
Month	  & 55 & 0 & 23 & 5 &  3 & 63 & 81 & 0.060 & 0.716 & 0.921 \\
Date	  & 49 & 6 & 23 & 5 &  3 & 63 & 81 & 0.060 & 0.642 & 0.825 \\
DayofWeek & 52 & 3 & 23 & 5 &  3 & 63 & 81 & 0.085 & 0.679 & 0.873 \\
HourMin	  & 34 & 3 &  7 & 6 & 36 & 79 & 80 & 0.852 & 0.875 & 0.886 \\
TimeDay	  & 18 & 8 & 31 & 2 & 27 & 55 & 84 & 0.354 & 0.536 & 0.818 \\
\hline 
end &  &  &  &  &  &  &  &  &  & \\ 
Month	  & 55	& 0 & 23 & 5 &  3 & 63 & 81 & 0.060 & 0.716 & 0.921 \\
Date	  & 49  & 6 & 23 & 5 &  3 & 63 & 81 & 0.060 & 0.642 & 0.825 \\
DayofWeek & 52  & 3 & 23 & 5 &  3 & 63 & 81 & 0.060 & 0.679 & 0.873 \\
HourMin   & 28  & 2 & 13 & 1 & 42 & 73 & 85 & 0.795 & 0.824 & 0.959 \\
TimeDay	  & 9   & 2 & 32 & 5 & 38 & 54 & 81 & 0.482 & 0.580 & 0.870 \\
\hline 
{\bf Overall} & 401 & 33 & 221 & 44 & 161 & 639	& 816 & 0.286 & {\bf 0.689} & {\bf 0.879} \\
\hline 
\end{tabular}
\caption{Evaluation of System on NMSU Test Data}
\label{nmsu-eval}
\end{center}
\end{table*}
}

The baseline accuracies for the test data sets are shown 
in Table \ref{lower-bounds}. These values were derived by disabling
all the rules and evaluating the input itself (after
performing normalization, so that the evaluation software could be applied).
Since null values are the
most frequent for all fields, this is equivalent to using a
naive algorithm that selects the most frequent value for each
field.  
Note that in Tables \ref{cmu-eval} and \ref{nmsu-eval},
the baseline accuracies for the end {\it month}, {\it date}, 
and {\it day of week} fields
are quite low because the coding convention calls for filling in
these fields, even though they are not usually explicitly
specified. In this case, an alternative baseline would have been to use the
corresponding starting field. This has not been calculated, but the
results can be approximated by using the baseline figures for the
starting fields.

The rightmost column of Table \ref{lower-bounds}
shows that there is a small
amount of error in the input representation.  This figure
is 1 minus the precision of the input representation (after
normalization).  Note, however, that this is a close but not
exact measure of the error in the input, because
there are a few cases of the normalization process committing errors
and a few of it correcting errors.  
Recall that the input is ambiguous; the figures in Table
\ref{lower-bounds} are based on the system selecting the first ILT in
each case. Since the parser orders the ILTs based on a measure of
acceptability, this choice is likely to have the relevant temporal
information.

{
\begin{table*}
\begin{center}
\noindent
\begin{tabular}{|l|r|r|r|r|r||r|r|r|r|}
\hline
Set    & Cor & Inc & Mis & Ext & Nul & Act & Poss & {\bf Acc}  & Input Error \\
\hline 
cmu    &  84 &   6 & 360 &  10 & 190 & 290 & 640  & {\bf 0.428} & 0.055 \\
nmsu   &  65 &   3 & 587 &   4 & 171 & 243 & 826  & {\bf 0.286} & 0.029 \\
\hline
\end{tabular}
\caption{Baseline Figures for Both Test Sets}
\label{lower-bounds}
\end{center}
\end{table*}
}

The above results are for the system taking ambiguous semantic
representations as input.  To help isolate errors due to our
model, the
system was also evaluated on unambiguous, partially
corrected input for all the seen data (the test sets were retained as
unseen test data).  The input is only partially corrected because
some errors are not feasible to correct manually,
given the complexity of the input representation.

The overall results are shown in the Table \ref{all-eval}. The table
includes the results presented earlier in Tables 
\ref{cmu-eval} and \ref{nmsu-eval}, to facilitate comparison.
In the CMU data set, there are twelve dialogs in
the training data and three dialogs in a held out test set.  The
average length of each dialog is approximately 65 utterances.  In
the NMSU data set, there are four training dialogs
and three test dialogs.  
{
\begin{table*}
\begin{center}
\begin{tabular}{|l|l|l|r|c|r|r|} 
\hline
seen/   & cmu/ & ambiguous, uncorrected/ & Dialogs
& Utterances &
{\bf Acc} & {\bf Prec} \\
unseen  & nmsu & unambiguous, partially corrected &          &             &         &
\\
\hline
seen    & cmu & ambiguous, uncorrected   
& 12 & 659 & 0.883 & 0.918 \\  
seen    & cmu & unambiguous, partially corrected & 12 & 659 & 0.914 & 0.957 \\  
unseen  & cmu & ambiguous, uncorrected   & 3  & 193 & 0.809 & 0.916 \\  
\hline
seen    & nmsu & ambiguous, uncorrected   & 4 & 358 & 0.679 & 0.746 \\  
seen    & nmsu & unambiguous, partially corrected & 4 & 358 & 0.779 & 0.850 \\  
unseen  & nmsu & ambiguous, uncorrected  & 3 & 236 & 0.689 & 0.879 \\  
\hline 
\end{tabular}

\end{center}
\caption{Overall Results}
\label{all-eval}
\end{table*}
}

In both data sets, there are noticeable gains in performance on the
seen data going from ambiguous to unambiguous input, especially for
the NMSU data. Therefore, the semantic ambiguity and input errors contribute
significantly to the system's errors.

Some challenging characteristics of the seen NMSU data are
vast semantic ambiguity,
numbers mistaken by
the input parser for dates (for example, phone numbers are treated as
dates),
and the occurrences of subdialogs.

Most of the the system's errors on the unambiguous data
are due to parser error, errors in applying the rules, 
errors in mistaking anaphoric references for deictic references
(and vice versa), and errors in choosing the wrong anaphoric relation.
As will be shown in Section \ref{focusmodelanal}, our approach handles
focus effectively, so few errors can be 
attributed to the wrong entities being in focus.

\section{Other Work on Temporal Reference Resolution}
\label{otherwork}
To our knowledge, there are no other published results on unseen test
data of systems performing similar temporal reference resolution
tasks.  Ros\'{e} et al.~\citeyear[Enthusiast]{rose-etal95},
Alexandersson et al.~\citeyear[Verbmobil]{alexandersson-etal97}, and
Busemann et al.~\citeyear[Cosma]{busemann-etal97} describe other
recent natural language processing systems that resolve temporal
expressions in scheduling dialogs.  Ros\'e et al.~also address focus
issues; we compare our work to theirs in detail in Section
\ref{focusmodelanal}.  All of the systems share certain features, such
as the use of a calendar utility to calculate dates, a specificity
ordering of temporal components (such as in Figure \ref{specificity}),
and a record of the temporal context.

However, all of the other systems perform
temporal reference resolution as part of their overall processing, in
service of solving another problem such as speech act resolution.
None of them lays out a detailed approach or model for temporal reference
resolution, and none gives results of system performance on any
temporal interpretation tasks.

Kamp and Reyle \citeyear{kampreyle93} address representational and
processing issues in the interpretation of temporal expressions.
However, they do not implement their ideas or present the results of a
working system. They do not attempt coverage of a data set, or present
a comprehensive set of relations, as we do, but consider only specific
cases that are interesting for their Discourse Representation Theory.
In addition, they do not address the issues of discourse
structure and attentional state focused on here.  For example, they
recognize that references such as ``on Sunday'' may have to be
understood with respect to a frame of reference. 
But they do not address how the frame of
reference is chosen in context, so do not address the question
of what type of focus model is required.

Note that temporal reference resolution is a different problem
from tense and aspect interpretation in discourse 
\cite<as addressed in, for example,>{webber88,songcohen91,hwangschubert92,lascarides-etal92,kameyama-etal93}.   
These tasks are briefly reviewed here to
clarify the differences.
Temporal reference resolution is
determining
what time is being explicitly specified by noun phrases that are
temporal referring
expressions (e.g.,
``Monday'' resolved to {\it Monday 19 August}).  Tense and aspect
interpretation involves determining implicit information about the states
and events specified by verb phrases (e.g., that the kissing event
specified in ``He had kissed her'' happened before some reference
time in the past).  While it could aid
in performing temporal reference resolution, we are not addressing
tense and aspect interpretation itself.

Scheduling dialogs, or scheduling subdialogs of other kinds of
dialogs, predominantly employ the present and future tenses, 
due to the nature of
the task.  As discussed further below in Section \ref{focusmodelanal},
a primary way that tracking the tense and aspect would aid in
temporal reference resolution would be to recognize discourse segments
that depart from the scheduling dialog or subdialog.  In addition,
Kamp and Reyle \citeyear{kampreyle93} address some cases in which tense and aspect,
temporal nouns, and temporal adverbs interact to affect the temporal
interpretation.  We intend to pursue these ideas in future work.

\section{Analysis}
\label{analysis}
The implementation is an important proof of concept.
However, as discussed in Section \ref{results}, various kinds of errors are
reflected in the results; many are not directly related to discourse
processing or
temporal reference resolution.  Examples are (1) completely null inputs,
when the semantic parser or speech recognizer fails, (2) numbers mistaken
as dates, and (3) failures to recognize that a relation can
be established, due to a lack of specific domain knowledge.

To evaluate the algorithm itself, in this section we separately
evaluate the components of our method for temporal reference
resolution.  Sections \ref{focusmodelanal} and \ref{relanal} assess
the key contributions of this work: the focus model (in Section
\ref{focusmodelanal}) and the deictic and anaphoric relations
(in Section \ref{relanal}). These evaluations required us to perform
extensive additional manual annotation of the data.  In order to
preserve the test dialogs as unseen test data, these
annotations were performed on the training data only.  In Section
\ref{archeval}, we isolate the architectural components of our
algorithm, such as the certainty factor calculation and the critics, to
assess the effects they have on performance.

\subsection{Evaluation of the Focus Model}
\label{focusmodelanal}
The algorithm presented here does not include a mechanism for
recognizing the global structure of the discourse, such as in
the work of
Grosz and Sidner \citeyear{groszsidner86}, Mann and Thompson
\citeyear{mannthompson88}, Allen and Perrault
\citeyear{allenperrault80}, and in
descendent work.  Recently in the literature, Walker \citeyear{walker96} 
argues for a more linear-recency based model of attentional state
(though not that discourse structure need not be recognized), while
Ros\'{e} et al.~\citeyear{rose-etal95} argue for a more complex model of attentional
state than is represented in most current computational theories of
discourse.

Many theories that address how attentional state should be modeled
have the goal of performing intention recognition as well.
We investigate performing temporal reference resolution directly,
without also attempting to recognize discourse structure or
intentions.  We assess the challenges the
data present to our model when only this task is attempted.

The total number of Temporal Units and the number of them
specified by anaphoric noun phrases
in the two training data sets are given in
Figure \ref{refcounts}.\footnote{The anaphoric counts include the cases
in which both deictic and anaphoric interpretations yield the 
correct result.}
There are different units that
could be counted, from the number of temporal noun phrases to the
number of distinct times referred to in the dialog. Here, we count the
entities that must be resolved by a temporal reference
resolution algorithm, i.e., the number of
distinct temporal units specified in each sentence, summed over all
sentences.  Operationally, this is a count of Temporal
Units after the normalization phase, i.e., after Step 1 in Section
\ref{architecture}.  This is the unit considered in the remainder of
this paper.

\begin{figure}
\begin{center}
\begin{tabular}{|l|c|c|}
\hline
 & \# TUs  & \# TUs specified anaphorically \\
\hline
CMU    & 196  & 167 \\  
\hline
NMSU   &  96  &  71 \\  
\hline
Total  & 292  & 238 \\
\hline
\end{tabular}
\caption{Counts of Temporal Unit References in the Training Data}
\label{refcounts}
\end{center}
\end{figure}

\begin{figure}
\centerline{\psfig{figure=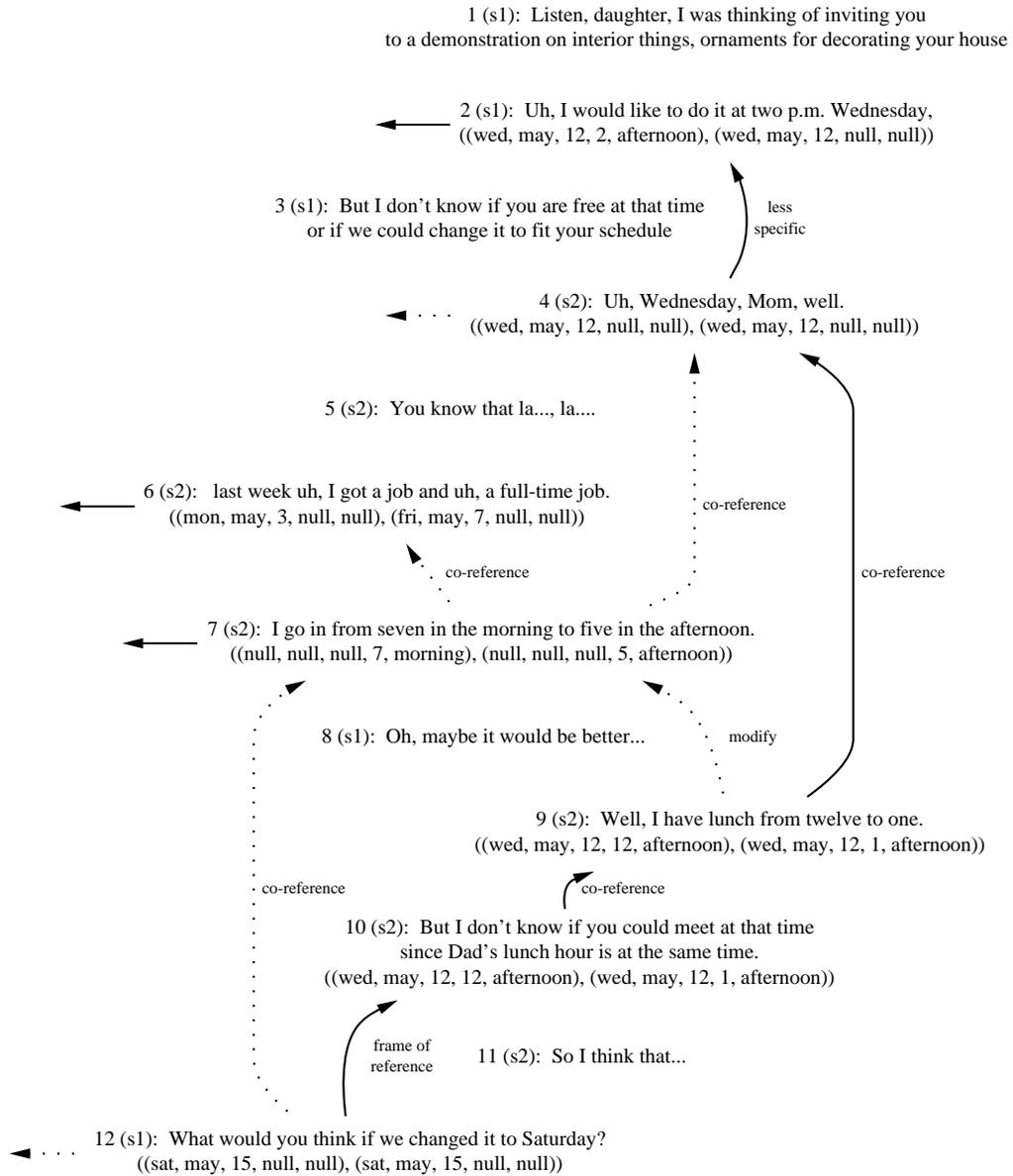,width=0.99\textwidth}}
\caption{Anaphoric Annotations of Part of NMSU Dialog 9.}
\label{anachains}
\end{figure}

To support the evaluation presented in this section, antecedent
information was manually annotated in the training data.  For each
Temporal Unit specified by an anaphoric noun phrase,
all of the antecedents that yield the correct
interpretation under one of the anaphoric relations were identified,
except that, if both $TU_i$ and $TU_j$ are appropriate antecedents,
and one is an antecedent of the other, only the more recent one is
included.  Thus, only the heads of the anaphoric chains existing at
that point in the dialog
are included.  In addition, {\it competitor} discourse entities 
were also identified,
i.e., previously mentioned Temporal Units
for which some relation could be established, but the
resulting interpretation would be incorrect.
Again, only Temporal Units at the head of an anaphoric 
chain were considered.
To illustrate these annotations,
Figure \ref{anachains} shows a graph depicting 
anaphoric chain annotations of an NMSU dialog (dialog 9).
In the figure, solid lines link the correct antecedents, dotted lines
show
competitors, and edges to nowhere indicate deictics.

\subsubsection{Cases in which the immediately preceding time is not an
appropriate antecedent}
\label{countback}

The main purpose of a focus model is to make an appropriate set of
discourse entities available as candidate antecedents at each point in
the discourse.  As described above in Section \ref{focusmodel}, Grosz
and Sidner's model captures situations in
which entities should not be available as candidate antecedents,
and  Ros\'e et al.~identify
situations in which
Grosz and Sidner's model may incorrectly eliminate entities from
consideration (i.e., dialogs with multiple threads).  
The potential challenge for a recency-based model like
ours is that entities may be available as candidate antecedents that
should not be.  An entity $E$ may occur to which an anaphoric relation
could be established, but an entity mentioned before $E$ is needed for
the correct interpretation. (From another perspective, $E$ yields the
wrong interpretation but cannot be ruled out as a possible
antecedent.)  To assess the magnitude of this problem for our method,
in this section we characterize the cases in which the most recent
entity is not an appropriate antecedent.

Before proceeding, we note that there is only one situation in which
our model incorrectly makes a needed entity {\it un}available.  Recall
from Section \ref{focusmodel} that, for a particular relation $R$,
only the most recent Temporal Unit for which $R$ can be established is
a candidate (call it $C$).  The problem arises when the correct
interpretation requires that that same relation $R$ be
established with an entity mentioned earlier than $C$.  This is a
problem because the earlier time is not a candidate.  If such cases
were to occur in the training data, they would have been found by the
analysis presented below.  However, none were found.

Based on the anaphoric chain annotations, we identified how far back
on the focus list one must go to find an antecedent that is
appropriate according to the model.  An antecedent is considered to be
appropriate according to the model if there exists a relation defined
in the model such that, when established between the current utterance
and the antecedent, it yields the correct interpretation.  Note that
we allow antecedents for which the anaphoric relation would be a
trivial extension of one of the relations explicitly defined in the
model.  For example, phrases such as ``after lunch'' should be treated
as if they are simple times of day under the {\it co-reference} and {\it
modify} anaphoric relations, but, as explicitly defined, those
relations do not cover such phrases.  For example, given {\it
Wednesday 14 April}, the reference ``after lunch'' should be
interpreted as {\it after lunch, Wednesday 14 April} under the {\it
co-reference} relation. 
Similarly, given {\it 10am, Wednesday, 14
April}, ``After lunch'' in ``After lunch would be better''
should be interpreted as {\it after lunch, Wednesday 14 April} under the
{\it modify} anaphoric relation.

The results are striking.  Between the two sets of training data,
there are only nine anaphoric temporal references for which the
immediately preceding Temporal Unit is not an appropriate antecedent,
3/167 = 1.8\% in the CMU data, and 6/71 = 8.4\% in the NMSU data.

\begin{figure}
\begin{center}
\centerline{
\psfig{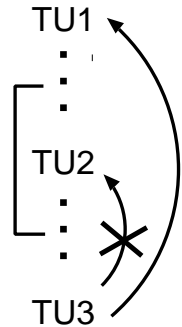}}
\caption{Structure Challenging the Recency Model.}
\label{struct}
\end{center}
\end{figure}

Figure \ref{struct} depicts the structure involved in all nine
cases. 
$TU_3$ represents the
anaphoric reference for which the immediately preceding Temporal Unit
is not an appropriate antecedent.  $TU_1$ represents the most recent
appropriate antecedent, and $TU_2$ represents the intervening
Temporal Unit or Units.  The ellipses represent any
intervening non-temporal
utterances.

Figure \ref{summaryfigure} characterizes the nine cases along a number
of dimensions.
To isolate the issues addressed, it was assumed in deriving these figures that the
dialog is correctly interpreted up to and including $TU_1$. 
\begin{figure}
\small
\begin{center}
\begin{tabular}{|c|c|c|c|c|c|c|}
\hline
& \multicolumn{1}{c|}{1}  &  \multicolumn{1}{c|}{2} &  \multicolumn{1}{c|}{3}  & \multicolumn{1}{c|}{4}
& \multicolumn{1}{c|}{5}  & \multicolumn{1}{c|}{6}  \\
\hline
& Distance to & Subdialog? & Type of $TU_2$ & $TU_2$  & $TU_2$ a & Potential \\
& most recent &            &             & Correct?    & Competitor?                   & Cumulative  \\
   & appropriate & & & & & Errors \\
   & antecedent & & & & & \\
\hline
1 (07-37)   & 2      &  No        & Anaphoric,  &  Yes   &   Yes   &   21 \\
CMU         &        &            & main task   &        &         & \\
\hline
2 (07-63) &  2          &  No        & Habitual &       No & Yes            &   0 \\
CMU       &             &            &          &          &                & \\ 
\hline
3 (15-31)   & 2           &  No        & Anaphoric, &   Yes         &   Yes              & 4 \\
CMU      &             &            & main task      &  & & \\
\hline
4 (08-57)  & 2          &   Yes       & Reference    & No   &  Yes & 2 minor \\
NMSU     &            &             & outside dialog &   & & \\
\hline
5 (08-66)  &  3         &   Yes       & 1 deictic   &  Yes  & Yes  & 10   \\
NMSU      &            &             & 1 habitual  &  No   & Yes  &  (worst case) \\
\hline
6 (09-39)  &  2          & No          & habitual    & No   &  No   & 0 \\
NMSU       &             &            &          &          &                & \\ 
\hline
7 (09-09)&   3          & Yes         & 1 deictic   & Yes  & Yes  &  4 \\
NMSU     &             &             & 1 habitual  & No   &      & (worst
case) \\
\hline
8 (09-45)  &  3    & Borderline   & both habitual  & No & Yes   & 6 \\
NMSU      &        &              &                &    &      & \\
\hline 
9 (10-55) & 3 & Borderline & both habitual & No & Yes &
3 \\
NMSU     &             &             & &   &      &  \\
\hline

\end{tabular}
\caption{Summary of Cases in Which Most Recent TU is not an
Appropriate Antecedent}
\label{summaryfigure}
\end{center}
\end{figure}

In three of the cases (rows 2, 4, and 9, labeled 07-63, 08-57, 10-55,
respectively), there is a correct deictic interpretation of $TU_3$
under our model, in addition to the correct (with antecedent $TU_1$)
and incorrect (with antecedent $TU_2$) anaphoric interpretations.

Column 1 of Figure \ref{summaryfigure} shows that,
in all three cases in the CMU data and in two cases in the NMSU data, the
second most recently mentioned Temporal Unit is an appropriate antecedent.
In the remaining four cases, the third most recently mentioned time is
appropriate.

In three of the cases, the references represented by $TU_2$ in Figure
\ref{struct} are in subdialogs off the main topic and scheduling task
(indicated as ``Yes'' in column 2).  All of these subdialogs are in
the NMSU data.
In four cases, the $TU_2$
references are in subsegments that are directly in service of the main
task (indicated as ``No'' in column 2), and in two cases, we judged
them to be borderline.

Column 3 characterizes the type of reference the $TU_2$ references are.
The two  marked ``Anaphoric, main task'' are specific references to times that
involve the main scheduling task.   The 
subdialog marked ``Reference
outside dialog'' (row 4, label 8-57) 
is shown in Figure \ref{row4}. 
\begin{figure}
\begin{center}
\begin{tabular}{|ll|}
\hline
\multicolumn{2}{|c|}{{\it Dialog Date:  Monday 10 May}}  \\
\hline
$TU_1$:  & It's just that $\dots$ this Thursday {\it [Thursday May 13]} is our second wedding
\\
         & anniversary and I don't know what to do.  \\
& $\langle$ 31 non-temporal utterances about what to cook $\rangle $ \\
& Did you go with my mother?  \\
$TU_2$: &  With my mother? Yes.  I went at around six in the morning. \\
&  Did you and Maura go for a walk? \\
& No, no we didn't. \\
& Hmmmmm.   We got lazy. \\
& Ah Claudia. \\
$TU_3$ & Well, yes.  Listen Lily.  What do you think if we see each
other on, \\ 
& on Thursday at six and I, at six? \\
\hline
\end{tabular}
\end{center}
\caption{Dialog Segment of the Case in Row 4 in Figure \ref{summaryfigure}}
\label{row4}
\end{figure}

The main topic of this dialog is a
party for the anniversary mentioned in $TU_1$.
The $TU_2$ reference, ``around six in the morning,'' involves the
participants' shared knowledge of an event that is not related to the
scheduling task.  The only interpretation possible in our model is six
in the morning on the day specified in the $TU_1$ reference, while in
fact the participants are referring to six in the morning on the dialog date. (There
is currently no coverage in our model for deictic references that mention
only a time of day.)  Thus, the interpretation of the $TU_2$ reference
is incorrect, as indicated in column 4.

Many of the $TU_2$ references are habitual (marked
``habitual'' in column 3 of Figure \ref{summaryfigure}). 
For example, the participants discuss their usual work schedules,
using utterances such as ``during the week I work from 3 to 6.''
Since there is no coverage of habituals in our model, the
interpretations
of all of the $TU_2$ habitual references are incorrect, as indicated
in column 4.  

We now turn to column 5, which asks a key 
question: is $TU_2$
a competitor?
$TU_2$ is a competitor if there is some
relation in the model that can be established between $TU_3$ and $TU_2$.
In the cases in which $TU_2$ represents multiple utterances
(namely, the fifth, seventh, eighth, and ninth rows of Figure \ref{summaryfigure}),
``yes'' is indicated in column 5 if an interpretation of the segment involving
both of the $TU_2$ references is possible.
Cumulative error (column 6) can be non-zero only if the entry in
column 5 is ``Yes'':  if the $TU_2$ references are
not competitors, they cannot be antecedents under our model, so
they cannot prevent $TU_3$ from being recognized as a correct antecedent.

It is important to note that the incorrect interpretation of
$TU_3$ and
the cumulative errors indicated in column
6 are only potential errors. In all cases in Figure
\ref{summaryfigure}, the correct
interpretation of $TU_3$ involving $TU_1$ is available as a possible
interpretation.  What is shown in column 6 is the number of
cumulative errors that would result if an interpretation involving
$TU_2$ were chosen over a correct interpretation involving $TU_1$.
In many cases, the system's answer is
correct because the (correct) $TU_3$--$TU_1$ interpretation involves
the {\it co-reference} anaphoric relation, while the (incorrect) $TU_3$--$TU_2$
interpretation involves the {\it frame of reference} anaphoric relation;
the certainty factor of the former is sufficiently
larger than that of the latter to overcome the distance-factor
penalty.  In addition, such interpretations often involve
large jumps forward in time, which are penalized by the 
critics. 

The worst case of cumulative error, row 1, is an example.
The segment is depicted
in Figure \ref{row1}.
\begin{figure}
\begin{center}
\begin{tabular}{ll}
\multicolumn{2}{l}{{\it Correct Interpretation of the $TU_1$ reference: Monday 22nd November}} \\
$TU_2$: & of December?  \\
$TU_3$: & of November.  \\ 
\end{tabular}
\end{center}
\caption{Dialog Segment of the Case in Row 1 in Figure \ref{summaryfigure}}
\label{row1}
\end{figure}
The incorrect interpretation involving $TU_2$
is November of the following year, 
calculated under the
{\it frame of reference} anaphoric relation.
The participants do not discuss the year, so the system
cannot recover.
Thus, a large amount of cumulative error
would result if that interpretation were chosen.  

\begin{figure}
\begin{center}
\centerline{
\psfig{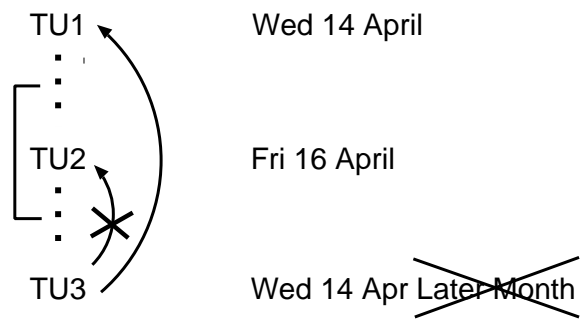}}
\caption{Structure of the Case in Row 3 of Figure \ref{summaryfigure}}
\label{row3}
\end{center}
\end{figure}
The segment corresponding to row 3 is similar.  Its structure 
is depicted in Figure \ref{row3}.
In
this passage, 
two days are mentioned in sequence, {\it Wednesday 14 April}
(the $TU_1$ reference) and {\it Friday 16 April} (the $TU_2$ reference).
Then, the day mentioned first---{\it Wednesday 14 April}---is referred to
again as ``Wednesday the 14th'' (the $TU_3$ reference).  There is no
relation in our model that enables the correct interpretation of
$TU_3$ to be obtained from $TU_2$.  
If $TU_2$ were taken to be the
antecedent of $TU_3$, the resulting incorrect interpretation would be
the next possible {\it Wednesday 14}, in a later month (possibly in a
later year), under the {\it frame of reference} anaphoric relation.
What is
required for the correct interpretation is the {\it co-reference} anaphoric
relation to be established between $TU_1$ and $TU_3$.
We saw exactly the same pattern above for the row 1
discourse segment, depicted in Figure \ref{row1}, except that in that
case a later month was calculated, rather than a later date.

It should be noted that,
if times rather than days or months were being discussed,
the correct interpretation for $TU_3$ could be obtained
from $TU_2$ under the {\it modify} anaphoric relation.  
A good example of this occurs in the corpus example in Figure
\ref{corpusexample}, repeated here as Figure \ref{repeat}.
\begin{figure}
\begin{center}
\begin{tabular}{|llll|}
\hline
\multicolumn{4}{|c|}{{\it Temporal context:
Tuesday 28 September}} \\
\hline
& s1& 1&   On Thursday I can only meet after two pm \\
&   &2&  From two to four \\
$TU_1$ &   &3&     Or two thirty to four thirty \\
$TU_2$ &   &4&    Or three to five \\
$TU_3$ & s2& 5&   Then how does from two thirty to  \\
&  &  & four thirty seem to you \\
&   &6&   On Thursday \\
& s1& 7&  Thursday the thirtieth of September \\
\hline
\end{tabular}
\caption{Corpus Example from Figure \ref{corpusexample}}
\label{repeat}
\end{center}
\end{figure}
The {\it modify} anaphoric relation enables $TU_2$ to be the antecedent
of $TU_3$.  The same would be true in the simpler
case of ``Two?  Or Three?  How about Two?''.
A promising future extension would be to 
develop a new {\it modify} anaphoric relation
for these cases.

Returning to column 6 of Figure \ref{summaryfigure},
note that two of the cumulative error figures are listed
as ``worst case.''  These are cases in which there are
two $TU_2$ references and there are many different possible
interpretations of the passage.

Notice that the second and fourth rows correspond to cases in which
$TU_2$ is a competitor, yet no significant potential cumulative error
results (the minor errors listed for row 4 are due to the relation
not fitting exactly, rather than an error from choosing the wrong
antecedent: {\it six in the morning} rather than {\it in the morning}
is placed into the high specificity fields).  In both of these cases,
the error corrects itself: $TU_1$ is incorrectly taken to be the
antecedent of $TU_2$, which is in turn incorrectly taken to be the
antecedent of $TU_3$.  But $TU_2$ in effect copies over the
information from $TU_1$ that is needed to interpret $TU_3$.  As a
result, the interpretation of $TU_3$ is correct.

In the cases for which there are only a few potential cumulative
errors, 
either a new, unambiguous time is soon introduced,
or a time being discussed before the offending
$TU_2$ reference is soon {\it re}introduced, getting things back on track.

An important discourse feature of the dialogs  is the degree of redundancy
of the times mentioned \cite{walker96}. This limits the ambiguity of the times
specified, and it also leads to a higher level of robustness, since
additional Temporal Units with the same time are placed on the focus
list and previously mentioned times are reintroduced. 
Table \ref{redundancy}
presents measures of redundancy.
The redundancy is broken down into the case
where redundant plus additional information is provided 
({\it Redundant}) versus the case where the temporal information
is just repeated ({\it Reiteration}). This shows that
roughly 27\% of the CMU utterances with temporal information contain
redundant temporal references, while 20\% of the NMSU ones do.

\begin{table*}
\begin{center}
\begin{tabular}{|c|c|c|c|c|} 
\hline 
Dialog Set & Temporal Utterances & Redundant & Reiteration & \% \\ 
\hline 
cmu     & 210            & 36        & 20          & 26.7 \\ 
nmsu    & 122            & 11        & 13          & 19.7 \\ 
\hline
\end{tabular}
\end{center}
\caption{Redundancy in the Training Dialogs}
\label{redundancy}
\end{table*}

In considering how the model could be improved, in addition to adding
a new {\it modify} anaphoric relation for cases such as those in
Figures \ref{row1} and \ref{row3}, habituals are clearly an area for
investigation.  Many of the offending references are habitual, and all
but one of the subdialogs and borderline subdialogs involve habituals.
In a departure from the algorithm, the system uses a simple heuristic
for ignoring subdialogs: a time is ignored if the utterance evoking it
is in the simple past or past perfect.  This prevents some of the
potential errors and suggests that changes in tense, aspect, and
modality are promising clues to explore for recognizing subsegments in
this kind of data \cite<see, for example,>{groszsidner86,nakhimovsky88}.
\subsubsection{Multiple threads}
\label{threadsanal}

Ros\'e et al.~\citeyear[p.~31]{rose-etal95} describe 
dialogs composed of multiple threads as ``negotiation dialogues in
which multiple propositions are negotiated in parallel.''
According to Ros\'e et al., dialogs with such multiple threads
pose challenges to a stack-based discourse
model on both the intentional and attentional levels.  They posit a
more complex representation of attentional state to meet these
challenges, and improve their results on speech act resolution in
a corpus of scheduling dialogs by using their model of attentional
state.\footnote{They do not report how many multiple thread
instances appear in their data.}

\begin{figure}
\begin{center}
\centerline{
\psfig{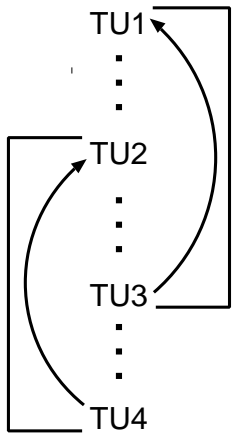}}
\caption{Temporal Multiple Thread Structure}
\label{struct-thread}
\end{center} 
\end{figure}
As discussed above, in this work we address only the attentional
level.  The relevant structure for temporal reference resolution,
abstracting from the examples given by Ros\'e et al.,
is shown in Figure \ref{struct-thread}.
There are four Temporal Units mentioned in the order $TU_1$, $TU_2$,
$TU_3$, and $TU_4$ (other times could be mentioned in between).  The
(attentional) multiple thread case is when $TU_1$ is required to be an
antecedent of $TU_3$, but $TU_2$ is also needed to interpret $TU_4$.
There are no realizations of this structure, in terms of
our model, in either the NMSU or CMU training
data set.

The case represented by row three in Figure \ref{summaryfigure}, whose
structure is depicted above in \ref{row3}, is the
instance in our data that is most closely related to
the situations addressed by Ros\'e et al.
This is a type of structure
that Grosz and Sidner's model
addresses, but it is not a multiple thread case, since $TU_2$ is not
needed to interpret a Temporal Unit mentioned after $TU_3$.

Ros\'e et al.'s examples of dialogs containing
multiple threads are shown in Figures \ref{roseex1} and
\ref{roseex2}, which are Ros\'e et al.'s Figures 1 and 2, respectively.
Figure \ref{roseex1} is an extended example, and
Figure \ref{roseex2} contains a simplified example which they
analyze in greater detail.  

\begin{figure}
\begin{center}
\begin{tabular}{|lll|}
\hline
\multicolumn{3}{|c|}{{\it Assumed Dialog Date:  Friday 11 April}}  \\
\hline
(1) & S1: & We need to set up a schedule for the meeting. \\
             & & \\
(2) & & How does your schedule look for next week? \\
(3) & S2: & Well, Monday and Tuesday both mornings are good.\\
(4) &     & Wednesday afternoon is good also. \\
(5) & S1: & It looks like it will have to be Thursday then. \\
(6) &     & Or Friday would also possibly work. \\
(7) &     & Do you have time between twelve and two on Thursday? \\
(8) &     & Or do you think sometime Friday afternoon you could meet?
\\
(9) & S2: & No. \\
(10) &    & Thursday I have a class. \\
(11) &    & And Friday is really tight for me. \\
(12) &    & How is the next week? \\
(13) &    & If all else fails there is always video conferencing. \\
(14) & S1: & Monday, Tuesday, and Wednesday I am out of town. \\
(15) &     & But Thursday and Friday are both good. \\
(16) &     & How about Thursday at twelve? \\
(17) & S2: & Sounds good. \\
& & \\
(18) & & See you then. \\
\hline
\end{tabular}

\caption{Example of Deliberating Over A Meeting Time}
\cite[p. 32]{rose-etal95}
\label{roseex1}
\end{center}
\end{figure}

The passage in Figure \ref{roseex1} 
would be processed by our algorithm as follows.
The dialog date is not given in \cite{rose-etal95}.  For concreteness,
let us suppose that the dialog date is {\it Friday 11 April}.  Then,
{\it next week} is {\it Monday 14 April} through {\it Friday 18 April} (the dialog
does not mention weekend days, so we exclude them for ease
of discussion).  
Utterance 2 is deictic, introducing 
{\it next week} into the discourse. 
Utterances 3-6 all have both deictic and anaphoric
readings, all of which yield the correct results.  

The deictic
relation for all of them is the {\it frame of reference}
deictic relation, under which the interpretations are
forward references from the dialog date:
\begin{center}
\begin{tabular}{cl}
Utterance & Deictic Interpretation \\
3 & Monday 14 April \& Tuesday 15 April \\
4 &  Wednesday 16 April \\
5 & Thursday 17 April \\
6 & Friday 18 April \\
\end{tabular}
\end{center}

The correct interpretations  of
(3)-(6) are also established with
the {\it co-reference} anaphoric relation, 
with antecedent {\it next week} in utterance 2:
they
each can be interpreted as specifying a more specific time than
{\it next week}, that is, as a particular day of {\it next week}. 

Finally, the 
{\it frame of reference} anaphoric relation yields the correct result for
``Tuesday'' in 
(3)\footnote{Recall that multiple Temporal Units specified in a single
utterance are added to the focus list in order of 
mention and treated as separate discourse entities.}
and for the times specified in utterances
(4)-(6).  The interpretation is the day calculated forward from the most
recently mentioned Temporal Unit:  

\begin{center}
\begin{tabular}{cll}
\multicolumn{1}{c}{Utterance} & \multicolumn{1}{c}{Antecedent} 
& \multicolumn{1}{c}{Interpretation} \\\\
3 & Monday 14 April, Utterance 3  & Tuesday 15 April \\
4 & Tuesday 15 April, Utterance 3 & Wednesday 16 April \\
5 & Wednesday 16 April, Utterance 4 &  Thursday 17 April \\
6 & Thursday 17 April, Utterance 5 & Friday 18 April \\
\end{tabular}
\end{center}

\noindent
Utterances (7) and (10) are potential challenges for our algorithm,
representing instances of the situation depicted in Figure
\ref{row3}:  {\it Thursday 24 April}
is a possible incorrect interpretation of ``Thursday'' in these
utterances, yielded by the 
{\it frame of reference} anaphoric relation.
The correct interpretation is also a candidate, yielded
by multiple relations: the {\it frame of reference}
deictic relation and the {\it co-reference} anaphoric relation,
with {\it Thursday 17 April} in utterance (5) as antecedent.  The relative
magnitude of the certainty factors of the {\it co-reference} and
{\it frame of reference} anaphoric relations 
means that the correct
interpretation is likely to be chosen in practice, as mentioned
in Section \ref{countback}.
If the incorrect interpretation
were chosen for utterances (7) and (10), 
then incorrect interpretations of ``Friday''
in each of (8) and (11) would be possible:  the Friday after the
incorrect date of {\it Thursday 24 April}, yielded by the {\it frame of reference}
anaphoric relation.
However, 
the correct interpretations would be possible too,
yielded by the {\it frame of reference} deictic relation and 
the {\it co-reference} anaphoric relation.

Utterances (12) through (16) have analogous interpretations,
except that the deictic interpretations yield incorrect
results (that is, due to utterance 12, ``How is the next week?'',
the days are actually of the week {\it Monday 21 April}
through {\it Friday 25 April}; the deictic interpretations are
of the week {\it Monday 14 April} through {\it Friday 18 April}).  Thus, there
are one correct and two incorrect interpretations for some
of the utterances, making it less likely in practice that
the correct interpretation would be chosen.  Note that,
generally speaking, 
which focus model is used does not directly address
the deictic/anaphoric ambiguity, so, for the purposes of this
section, the two parts of the dialog pose the same challenge
to the focus model.  

\begin{figure}
\begin{center}
\centerline{
\psfig{figure=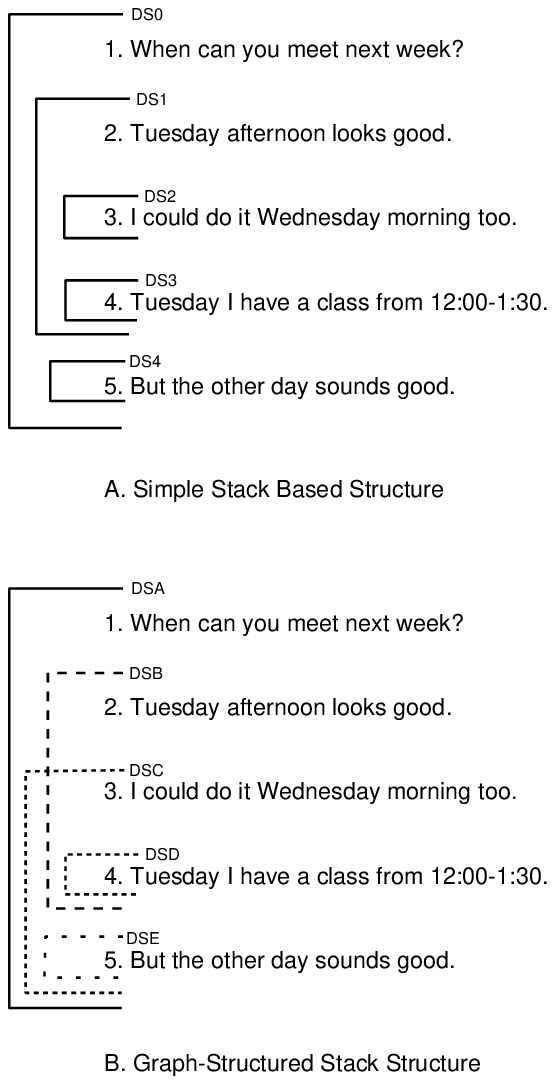,width=0.65\textwidth}}
\caption{Sample Analysis}
\cite[p. 33]{rose-etal95}
\label{roseex2}
\end{center}
\end{figure}

The dialog in Figure \ref{roseex2} is analogous.
However, ``The other day'' in (5) brings up other issues.  There is
a special case of the {\it co-reference} anaphoric relation for such expressions
(i.e., ``the other'' ``day''$\mid$``month''$\mid$``year''; see Anaphoric Rule 7 in
Online Appendix 1).
In this case, the second 
most recent day, month, or year, as appropriate, is the candidate 
antecedent.  Presumably, neither the most recently mentioned
day nor a day mentioned before two or more others 
would be referred to as ``the other day''; thus, we
anticipate that this is a good heuristic.
Nevertheless, if (5) explicitly mentioned ``Wednesday,''
our algorithm would have a correct and an incorrect interpretation
to choose between.

In summary, there were no instances of temporal multiple threads of
the type addressed by Ros\'e et al., either in the CMU training data
upon which the algorithm was developed, or in the NMSU training data
to which the algorithm was later applied.
If segments
such as those illustrated in Ros\'e et al.~were to appear, an
incorrect interpretation by our algorithm would be possible,
but, under our model, the
correct antecedent would also be available. 
For the examples they
present, the algorithm faces the same choice: establish a {\it co-reference}
relation to a time before the last one (the correct interpretation),
or establish a {\it frame of reference} relation with the immediately
preceding time (an incorrect interpretation).  If performing temporal
reference resolution is the goal, and if one is faced with an application in
which such temporal multiple threads do occur, our 
investigation of the problem suggests that this specific
situation 
should be investigated before assuming that a more
complex focus model is needed.  Adding a new {\it modify} anaphoric
relation could be investigated.
Or, as implemented in our system, a specific
preference could be defined for the {\it co-reference} relation over the
{\it frame of reference} relation when both are possible in a local context.
Statistical techniques could be used to establish
preferences appropriate for the particular application.   

The different findings between Ros\'e et al.~and our
work might be due to the fact that different
problems are being addressed.  Having no intentional state, our model
does not distinguish between times being negotiated and other times.  It is
possible that another structure is relevant for the intentional level.
Ros\'{e} et al.~do not specify whether or not this is so.  The
different findings may also be due to differences in the data: 
their protocol is like a radio conversation in which a
button must be pressed in order to transmit a message, and the
other participant cannot transmit a message until the speaker
releases the button.
This results in less
dynamic interaction and longer turns \cite{villa94}.
In the dialogs used here, the participants have free control
over turn-taking.
\subsection{Coverage and Ambiguity of the Relations Defined in the
Model}
\label{relanal}
A question naturally arises from the evaluation presented in the
previous section:  
in using a less complex focus model, have we merely ``pushed
aside'' the ambiguity into the set of deictic and anaphoric 
relations?  In this section,
we assess the ambiguity of the anaphoric relations for the NMSU
and CMU training sets.
This section also presents other evaluations of the
relations, including an assessment of their coverage, redundancy,
how often they are correct, and how often they are applicable.

The evaluations presented in this section required detailed, 
time-consuming manual annotations.  The system's annotations would not
suffice, because the implementation does not perfectly recognize when a
rule is applicable.  A sample of four randomly
selected dialogs in the CMU training set and the four
dialogs in the NMSU training set were annotated.

The counts derived from the manual annotations for this section are defined
below.  Because this section focuses on the relations, we
consider them at the more specific level of the deictic and anaphoric rules 
presented in Online Appendix 1. In addition, we do
not allow trivial extensions of the relations, as we did in the evaluation
of the focus model (Section \ref{focusmodelanal}).  The criterion for
correctness in this section is the same as for the evaluation of the system:
a field-by-field exact match with the manually annotated correct interpretations.
There is one exception.  The starting and end {\it time of day} fields are ignored,
since these are known weaknesses of the rules, and they represent a relatively
minor proportion of the overall temporal interpretation.

The following were derived from manual annotations.
\begin{itemize}
\item
{\it TimeRefs:} the number of distinct times referred to in each sentence,
summed over all sentences.
\item
{\it TimeRefsC:} The number of {\it TimeRefs} 
for which a correct interpretation
is available under our model (whether or not an incorrect interpretation
is also possible).
\item
{\it Interp:} The number of interpretations possible under the model.  For
the current Temporal Unit, 
there is one {\it Interp} for every rule that can be applied.
\item
{\it CorrI:}  The number of {\it Interps} that are correct, 
where correctness is
defined as an exact match with the manually annotated correct
interpretation, except that the starting and end {\it time of day}
fields are ignored.
\item
{\it IncI:}  The number of incorrect {\it Interps} (i.e., {\it Interp}
= {\it IncI} +
{\it CorrI}).
\item
{\it DiffI:}   The number of different interpretations
\item
{\it DiffICorr:}  The number of different interpretations, excluding
interpretations of Temporal Units 
for which there is not a correct interpretation
under our model.
\end{itemize} 

The values for each data set, together with coverage and ambiguity evaluations,
are presented in Table \ref{covambtable}.
\begin{table}
\begin{center}
\begin{tabular}{|c|c|c|c|c|c|c|}
\hline
\multicolumn{7}{|c|}{CMU Training Set}  \\
\multicolumn{7}{|c|}{4 randomly selected dialogs}  \\
\hline
TimeRefs &  TimeRefsC &  Interp &  CorrI &  IncI &   DiffI &  DiffICorr \\
78      &   74       &  165    &  142   &   23  &    91   &   85 \\
\hline
\end{tabular}
\end{center}
\vspace*{2mm}
{\bf Coverage} (TimeRefsC / TimeRefs) = {\bf 95\%} \\
{\bf Ambiguity} (DiffICorr / TimeRefsC) = {\bf 1.15} \\
Overall Ambiguity (DiffI / TimeRefs) = 1.17 \\
Rule Redundancy (CorrI / TimeRefsC) = 142/74 = 1.92 \% \\

\begin{center}
\begin{tabular}{|c|c|c|c|c|c|c|}
\hline
\multicolumn{7}{|c|}{NMSU Training Set}  \\
\multicolumn{7}{|c|}{4 dialogs}  \\
\hline
TimeRefs &  TimeRefsC &  Interp &  CorrI &  IncI &   DiffI &  DiffICorr \\
   98   &   83       &   210   &   154   &  56   &   129  &   106  \\ 
\hline
\end{tabular}
\end{center}
\vspace*{2mm}
{\bf Coverage} (TimeRefsC / TimeRefs) = {\bf 85\%} \\
{\bf Ambiguity} (DiffICorr / TimeRefsC) = {\bf 1.28} \\
Overall Ambiguity (DiffI / TimeRefs) = 1.32 \\
Rule Redundancy (CorrI / TimeRefsC) = 154 / 83 = 1.86 \% \\
\caption{Coverage and Ambiguity}
\label{covambtable}
\end{table}

The ambiguity for both data sets is very low.  The {\it Ambiguity} figure
in Table \ref{covambtable} represents the average number of interpretations
per temporal reference, considering only those for which the correct interpretation
is possible (i.e., it is $DiffICorr$ / $TimeRefsC$).  The table also shows the ambiguity
when all temporal references are included (i.e., $DiffI$ / $TimeRefs$). 
As can be seen from the table, the average ambiguity in both
data sets is much less than two interpretations per utterance.

The coverage of the relations can be evaluated as ($TimeRefsC$ /
$TimeRefs$), the percentage of temporal references for which at least
one rule yields the correct interpretation.  While the coverage of the
NMSU data set, 85\%, is not perfect, 
it is good, considering that the system was not developed on the
NMSU data.

The data also show that there is often more than one way to achieve
the correct interpretation.  This is another type of redundancy:
redundancy of the data with respect to the model.  It is calculated in
Table \ref{covambtable} as ($CorrI$ / $TimeRefsC$), that is, the number of
correct interpretations over the number of temporal references that
have a correct interpretation.  For both data sets, there are, on
average,
roughly two different ways to achieve the correct interpretation.

Table \ref{relaccman} shows the number of times each rule applies
in total (column 3) and the number of times each rule is correct
(column 2),
according to our manual annotations.  Column 4 shows the accuracies
of the rules, i.e., 
(column 2 / column 3).
The rule labels are
the ones used in Online Appendix 1 to identify the rules.  

\begin{table}
\begin{center}
\begin{tabular}{|cccc|}
\hline
\multicolumn{4}{|c|}{CMU Training Set}  \\
\multicolumn{4}{|c|}{4 randomly selected dialogs} \\
\hline
Rule  &                    Correct &   Total &   Accuracy \\
\hline
D1     &        4 &     4   &    1.00  \\ 
\hline
D2i    &  0   &   0    &   0.00 \\
\hline
{\bf D2ii}      &  {\bf 35}  &   {\bf 40}   &   {\bf 0.88} \\
\multicolumn{4}{|l|}{\hspace*{2mm}{\it a frame-of-reference deictic
relation}} \\
\hline
D3       &  1   &   2    &   0.50 \\
\hline
D4        &  0   &   0    &   0.00 \\
\hline
D5        &  0   &   0    &   0.00 \\
\hline
D6          &  2   &   2    &   1.00 \\
\hline
{\bf A1}       &  {\bf 45}  &   {\bf 51}   &   {\bf 0.88} \\
\multicolumn{4}{|l|}{\hspace*{2mm}{\it a co-reference anaphoric
relation}} \\
\hline
A2       &  0   &   0    &   0.00 \\
\hline
A3i        &  1   &   1    &   1.00 \\
\hline
{\bf A3ii}           &  {\bf 35}  &   {\bf 37}   &   {\bf 0.95} \\
\multicolumn{4}{|l|}{\hspace*{2mm}{\it a frame-of-reference anaphoric
rel.}} \\
\hline
{\bf A4}        &  {\bf 14}  &   {\bf 18}   &   {\bf 0.78} \\
\multicolumn{4}{|l|}{\hspace*{2mm}{\it a modify anaphoric
relation}} \\
\hline
A5       &  0   &   0    &   0.00 \\
\hline
A6i       &  2   &   2    &   1.00 \\
\hline
A6ii  &  1   &   1    &   1.00 \\
\hline
A7       &  0   &   1    &   0.00 \\
\hline
A8      &  0   &   0    &   0.00 \\
\hline
\end{tabular}
\hspace*{0.5in}
\begin{tabular}{|cccc|}
\hline
\multicolumn{4}{|c|}{NMSU Training Set} \\
\multicolumn{4}{|c|}{4 dialogs} \\
\hline
Rule                  &   Correct &   Total&   Accuracy \\
\hline
D1           &  4   &   4    &   1.00 \\
\hline
D2i    &  0   &   0    &   0.00 \\
\hline
{\bf D2ii}       &  {\bf 24}  &   {\bf 36}   &   {\bf 0.67} \\
\multicolumn{4}{|l|}{\hspace*{2mm}{\it a frame-of-reference deictic
relation}} \\
\hline
D3       &  6   &   9    &   0.67 \\
\hline
D4        &  0   &   1    &   0.00 \\
\hline
D5        &  0   &   0    &   0.00 \\
\hline
D6          &  0   &   0    &   0.00 \\
\hline
{\bf A1}       &  {\bf 57}  &   {\bf 68}   &   {\bf 0.84} \\
\multicolumn{4}{|l|}{\hspace*{2mm}{\it a co-reference anaphoric
relation}} \\
\hline
A2       &  5   &   5    &   1.00 \\
\hline
A3i        &  0   &   0    &   0.00 \\
\hline
{\bf A3ii}           &  {\bf 21}  &   {\bf 32}   &   {\bf 0.66} \\
\multicolumn{4}{|l|}{\hspace*{2mm}{\it a frame-of-reference anaphoric
rel.}} \\
\hline
{\bf A4}        &  {\bf 27}  &   {\bf 37}   &   {\bf 0.73} \\
\multicolumn{4}{|l|}{\hspace*{2mm}{\it a modify anaphoric
relation}} \\
\hline
A5      &  0   &   1    &   0.00 \\
\hline
A6i       &  7   &   9    &   0.78 \\
\hline
A6ii   &  0   &   0    &   0.00 \\
\hline
A7       &  0   &   0    &   0.00 \\
\hline
A8      &  0   &   0    &   0.00 \\
\hline
\end{tabular}
\caption{Rule Applicability Based on Manual Annotations} 
\label{relaccman}
\end{center}
\end{table}

The same four rules are responsible for the majority of
applications in both data sets, the ones labeled
{\it D2ii}, {\it A1}, {\it A3ii}, and {\it A4}.  The first
is an instance of the {\it frame of reference} deictic relation,
the second is an instance of the {\it co-reference} anaphoric relation,
the third is an instance of the {\it frame of reference} anaphoric
relation, and the fourth is an instance of the {\it modify} anaphoric
relation.  

How often the system considers and actually uses each rule is shown in
Table \ref{sysrels}.  Specifically, 
the column labeled {\it Fires} shows
how often each rule applies, and the column labeled {\it Used} shows
how often each rule is used to form the final interpretation.  To help
isolate the accuracies of the rules, these experiments were performed
on unambiguous data.  
Comparing this table with Table
\ref{relaccman}, we see that
the same four rules shown to be the most important by the manual annotations
are also responsible for the
majority of the system's interpretations.  This holds for both the CMU
and NMSU data sets.

\begin{table}
\begin{center}
\begin{tabular}{|lrr|} 
\hline 
\multicolumn{3}{|c|}{CMU data set} \\
\hline 
\multicolumn{1}{|c}{Name} &
\multicolumn{1}{c}{Used} &
\multicolumn{1}{r|}{Fires} \\
\hline 
D1	 & 16	 & 16  \\
\hline 
D2i	 & 1	 & 3   \\
\hline 
{\bf D2ii}	 & {\bf 78}	 & {\bf 90}  \\
\multicolumn{3}{|l|}{\hspace*{2mm}{\it a frame-of-reference deictic
relation}} \\
\hline 
D3	 & 5	 & 5   \\
\hline 
D4	 & 9	 & 9   \\
\hline 
D5	 & 0	 & 1   \\ 
\hline 
D6	 & 2	 & 2   \\ 
\hline 
{\bf A1}	 & {\bf 95}	 & {\bf 110} \\
\multicolumn{3}{|l|}{\hspace*{2mm}{\it a co-reference anaphoric
relation}} \\
\hline 
A2	 & 2	 & 24  \\
\hline 
A3i	 & 1	 & 1   \\
\hline 
{\bf A3ii}	 & {\bf 72}	 & {\bf 86}  \\
\multicolumn{3}{|l|}{\hspace*{2mm}{\it a frame-of-reference anaphoric
rel.}} \\
\hline 
{\bf A4}	 & {\bf 45}	 & {\bf 80}  \\
\multicolumn{3}{|l|}{\hspace*{2mm}{\it a modify anaphoric
relation}} \\
\hline 
A5	 & 4	 & 5   \\
\hline 
A6i	 & 10	 & 10  \\
\hline 
A6ii	 & 0	 & 0   \\
\hline 
A7	 & 0	 & 0   \\
\hline 
A8	 & 1	 & 1   \\
\hline 
\end{tabular}
\hspace*{0.5in}
\begin{tabular}{|lrr|} 
\hline 
\multicolumn{3}{|c|}{NMSU data set} \\
\hline 
\multicolumn{1}{|c}{Name} &
\multicolumn{1}{c}{Used} &
\multicolumn{1}{r|}{Fires} \\
\hline 
D1	& 4	& 4  \\
\hline 
D2i	& 2	& 2  \\
\hline 
{\bf D2ii}	& {\bf 20}	& {\bf 31} \\
\multicolumn{3}{|l|}{\hspace*{2mm}{\it a frame-of-reference deictic
relation}} \\
\hline 
D3	& 2	& 3  \\
\hline 
D4	& 0	& 0  \\
\hline 
D5	& 0	& 0  \\
\hline 
D6	& 0	& 0  \\
\hline 
{\bf A1}	& {\bf 46}	& {\bf 65} \\
\multicolumn{3}{|l|}{\hspace*{2mm}{\it a co-reference anaphoric
relation}} \\
\hline 
A2	& 6	& 12 \\
\hline 
A3i	& 0	& 2  \\
\hline 
{\bf A3ii}	& {\bf 18}	& {\bf 27} \\
\multicolumn{3}{|l|}{\hspace*{2mm}{\it a frame-of-reference anaphoric
rel.}} \\
\hline 
{\bf A4}	& {\bf 24}	& {\bf 42} \\
\multicolumn{3}{|l|}{\hspace*{2mm}{\it a modify anaphoric
relation}} \\
\hline 
A5	& 3	& 5  \\
\hline 
A6i	& 6	& 8  \\
\hline 
A6ii	& 0	& 0  \\
\hline 
A7	& 0	& 0  \\
\hline 
A8	& 0	& 0  \\
\hline 
\end{tabular}
\caption{Rule Activation by the System on Unambiguous Data}
\label{sysrels}
\end{center}
\end{table}

\subsection{Evaluation of the Architectural Components}
\label{archeval}

In this section, we evaluate the architectural components of our
algorithm using degradation (ablation) studies.
We perform experiments without each component in turn, and
then with none of them, to observe the impact on the system's
performance.  Such studies have been useful in developing practical
methods for other kinds of anaphora resolution as well \cite<see, for example,>{mitkovstys97}. 
Specifically, an experiment was performed testing each
of the following variations.

\begin{enumerate}

\item
\label{cf1}
The certainty factors of all of the rules are set to 1.

Recall that all rules are applied to each utterance, and each rule
that matches produces a Partial-Augmented-ILT (which is assigned
the certainty factor of the rule).  All maximal mergings of the
Partial-Augmented-ILTs are then formed, to create a set of
Augmented-ILTs.  Then, the final interpretation of the utterance is chosen
from among the set of Augmented-ILTs.  The certainty factor of each
Augmented-ILT is the sum of the certainty factors of the
Partial-Augmented-ILTs composing it.  Thus, setting the certainty
factors to 1 implements the scheme in which the more partial results
are merged into an interpretation, the higher the overall certainty
factor of that interpretation. In other words, this scheme favors the
Augmented-ILT resulting from the greatest number of rule applications.

\item
\label{cf0}
The certainty factors of all of the rules are set to 0.  

This scheme is essentially random selection among the Augmented-ILTs that make
sense according to the critics.  If the critics did not exist, then
setting the rule certainty factors to 0 would result in random
selection.  With the critics, any Augmented-ILTs to which the critics
apply are excluded from consideration, because the critics will lower
their certainty factors to negative numbers.  

\item
\label{nomerge}
No merging of the rule results is performed.

That is, the Partial-Augmented-ILTs are not merged prior to selection
of the final Augmented-ILT. The effect of this is that the result of
one single rule is chosen to be the final interpretation.

\item
\label{nocritics}
The critics are not used.

\item
\label{nodistfacts}
The distance factors are not used.

In this case, the certainty factors for rules that access the focus
list are not adjusted based on how far back the chosen focus list item
is.

\item
\label{combo}
All variations are applied, excluding case \ref{cf0}.

Specifically, neither the critics nor the distance factors are used,
no merging of partial results is performed, and the rules are all
given the same certainty factor (namely, 1).
\end{enumerate}

Table \ref{variations-eval} shows the results for each variation when
run over the unambiguous but uncorrected CMU training data. For
comparison, the first row shows the results for the system as
normally configured. As with the previous evaluations, accuracy is the
percentage of the correct answers the system produces, while precision
is the percentage of the system's answers that are correct.

\begin{table}

\begin{center}
\begin{tabular}{|l|l|l|l|l|l||l|l|l|l|}
\hline
Variation       & Cor   & Inc   & Mis   & Ext  & Nul   & Act & Poss & {\bf Acc} & {\bf Prec} \\
\hline
{\bf system as is}& 1283 &  44	& 112	&  37	& 574 & 1938 & 2013 & 0.923 & 0.958 \\
all CFs 1.0     & 1261	&  77	& 101	&  50	& 561 & 1949 & 2000 & 0.911 & 0.935 \\
 all CFs 0.0    & 1202	& 118	& 119	&  49	& 562 & 1931 & 2001 & 0.882 & 0.914 \\
-critics        & 1228	& 104	& 107	& 354	& 667 & 2353 & 2106 & 0.900 & 0.805 \\
-dist. factors  & 1265	&  52	& 122	&  50	& 591 & 1958 & 2030 & 0.914 & 0.948 \\
-merge          & 1277	&  46	& 116	&  54	& 577 & 1954 & 2016 & 0.920 & 0.949 \\
combo           & 1270	&  53	& 116	&  67	& 594 & 1984 & 2033 & 0.917 & 0.940 \\
\hline
\end{tabular}
\end{center}

\begin{center}
\begin{tabular}{ll}
{\bf Legend}   & \\
Cor(rect):     & System and key agree on non-null value \\
Inc(orrect):   & System and key differ on non-null value \\
Mis(sing):     & System has null value for non-null key \\
Ext(ra):    & System has non-null value for null key \\
Nul(l):        & Both system and key give null answer \\
\\

Poss(ible): & Correct + Incorrect + Missing + Null \\
Act(ual): & Correct + Incorrect + Extra + Null \\
Base(line)Acc(uracy): & Baseline accuracy (input used as is) \\
Acc(uracy): & \% Key values matched correctly
 ((Correct + Null)/Possible) \\
Prec(ision):  & \% System answers matching the key
 ((Correct + Null)/Actual) \\

\end{tabular}
\end{center}
\caption{Evaluation of the Variations on CMU Unambiguous/Uncorrected Data}
\label{variations-eval}
\end{table}
\noindent

Only two of the differences are
statistically significant ($p \le 0.05$), namely, the precision of the
system's performance when the critics are not used, and the accuracy
of the system's performance when all of the certainty factors are 0. The
significance analysis was performed using paired t-tests comparing the results
for each variation with the results for the system as normally configured.

The performance difference when the critics are not used is due to
extraneous alternatives that the critics would have weeded out.  
The drop in accuracy when the certainty factors are all 0 shows that
the certainty factors have some effect. 
Experimenting with 
statistical methods to derive them
would likely lead to further improvement.

The remaining figures are
all only
slightly lower than those for the full system, and are all 
much higher than the baseline accuracies.

It is interesting to note that the unimportance of the distance
factors 
(variation 5) is consistent with the
findings presented in Section \ref{focusmodelanal}
that the last mentioned time is an acceptable antecedent in the vast
majority of cases.  Otherwise, we might have expected to see an
improvement in variation \ref{nodistfacts}, since the distance
factors penalize going further back on the focus list.

\section{Conclusions}
\label{conclusions}
Scheduling dialogs, during which people negotiate the times of
appointments, are common in everyday life.  This paper reports the
results of an in-depth empirical investigation of resolving explicit
temporal references in scheduling dialogs.  There are four basic
phases of this work: data annotation, model development, 
system
implementation and evaluation, and model evaluation and analysis.  The
system and model were developed primarily on one set of data (the CMU
dialogs), and then applied later to a much more complex set of data
(the
NMSU dialogs), to assess the generalizability of the model for the
task being performed.  Many different types of empirical methods were
applied to both data sets to pinpoint the strengths and weaknesses of
the approach.

In the data annotation phase, detailed coding instructions were
developed and an intercoder reliability study involving naive subjects
was performed. The results of the study are very good, supporting the
viability of the instructions and annotations.  During the model
development phase, we performed an iterative process of implementing a
proposed set of anaphoric and deictic relations and then refining them
based on system performance (on the CMU training data), until we
settled on the set presented here.  We also developed our focus model
during this phase.  The question of what type of focus model is
required for various tasks is a question of ongoing importance in the
literature.  It appeared from our initial observations of the data
that, contrary to what we expected, a recency-based focus model might
be adequate.  To test this hypothesis, we made the strategic
decision to limit ourselves to a recency-based model, rather than
build some kind of hybrid model whose success or failure would not
have told us as much.  

During system implementation and evaluation,
a system implementing the model was implemented and
evaluated on unseen test data, using a challenging
field-by-field comparison of system and human answers.  To be
considered the right answer, the information must not only be correct,
but must also be included in the correct field of the output
representation.  Taking as input the ambiguous output of a semantic
grammar, the system achieves an overall accuracy of 81\% on unseen
CMU test data, a large improvement over the baseline accuracy of
43\%.  
On an unseen test set from the more complex NMSU data,
the results are
very respectable: an overall accuracy of 69\%, with a
much lower baseline accuracy of 29\%.  
This also shows the
robustness of the CMU semantic parser \cite{lavietomita93,levin-etal95}, which was given the NMSU dialogs as input without being
modified in any way to handle them.

The implementation is an important proof of concept.  However, it is
not a direct evaluation of the model, because there are errors
due to factors we do not focus on in this work.
Some of the error is
simply due to utterance components being outside the coverage of the CMU
parser,
or having high semantic ambiguity.
The only information we use to perform semantic disambiguation is the
temporal context.  The Enthusiast researchers have already developed
better techniques for resolving the semantic ambiguity in these
dialogs \cite{shum-etal94}, which could be used to improve performance.

Thus, in the model evaluation and analysis phase, we performed
extensive additional evaluation of the algorithm itself.  We focus on
the relations and the focus model, because they are the main contributions
of this work.  Our degradation studies support this, as
they show that the other aspects of the algorithm, such as the
distance factors and merging process, are responsible for little
of the system's success (see Section \ref{archeval}).

Our evaluations show the strength of the focus model for the task, not
only for the CMU data on which it was developed, but also for the more
complex NMSU data.  While the NMSU data is more
complex, there are few cases in which the last mentioned
time is not an appropriate antecedent, highlighting the importance of
recency \cite{walker96}; see Section \ref{focusmodelanal}.  We
characterized those cases along a number of dimensions, to identify the
particular types of challenges they pose (see Figure
\ref{summaryfigure}).  

In order to compare our work to that of others, we formally defined 
subdialogs
and the multiple thread structures addressed 
by Ros\'e et al.~\citeyear{rose-etal95} 
with
respect to our model and the specific problem of temporal reference
resolution.  An interesting finding is that, while subdialogs of the
types addressed by Grosz and Sidner \citeyear{groszsidner86} were found in the data, no
cases of multiple threads were found.  That is, some subdialogs, all
in the NMSU data, mention times that potentially interfere with the
correct antecedent.  But in none of these cases would subsequent
errors result if, upon exiting the subdialog, the offending
information were popped off a discourse stack or otherwise made
inaccessible.  Changes in tense, aspect, and modality are promising
clues for recognizing subdialogs in this data, which we plan to
explore in future work.

To assess whether or not using a simpler focus model requires one to
use a highly ambiguous set of relations, we performed a separate
evaluation of the relations, based on detailed, manual
annotations of a set of dialogs.  The ambiguity of the relations for
both data sets is very low, and the coverage is good (see Table
\ref{covambtable}).  In a comparison of system and human annotations, the
same four rules identified to be most important in the manual
annotations are responsible for the majority of the system's
interpretations for both data sets (see Tables \ref{relaccman} and
\ref{sysrels}),
suggesting that the system is a good implementation of the model.

Recently, many in computational discourse processing have turned to
empirical studies of discourse, with a goal to develop general
theories by analyzing specific discourse phenomena and systems that
process them \cite{walkermoore97}.  We contribute to this general
enterprise.   We performed many different evaluations, on the CMU data
upon which the model was developed, and on the more
complex NMSU data.  The task
and model components were explicitly specified to facilitate evaluation and
comparison.  Each evaluation is directed toward answering a
particular question; together, the evaluations paint an overall
picture of the difficulty of the task and of the success of the
proposed model.

As a contribution of this work, we have made available on the project web
page the coding instructions, the NMSU dialogs, and the various kinds
of manual annotations we performed.  

\section{Acknowledgements}
This research was supported in part by the Department of
Defense under grant number 0-94-10.   
A number of people contributed to this work.  We want to
especially thank David Farwell, Daniel Villa, Carol Van Ess-Dykema, 
Karen Payne,
Robert Sinclair, Rocio Guill\'{e}n,  David Zarazua,
Rebecca Bruce, Gezina Stein, Tom Herndon, and
the project members of Enthusiast at CMU, 
whose cooperation greatly aided our project.
We wholeheartedly thank the anonymous reviewers, whose
comments and criticisms were very helpful.
We also thank Esther Steiner, Philip Bernick, and Julie England
for participating in the intercoder reliability study,
and Linda Fresques for proofreading the paper.

\vskip 0.2in

\bibliographystyle{theapa}
\bibliography{journal98}

\end{document}


\begin{center}
{\large \bf Online Appendix to {\em An Empirical Approach to Temporal
Reference Resolution}} \\
\noindent
\begin{tabular}{l} 
Janyce M. Wiebe, Thomas P. O'Hara, Thorsten~\"{O}hrstr\"{o}m-Sandgren \& Kenneth J. McKeever \\
\end{tabular}
\end{center}

\restylefloat{figure}
\restylefloat{table}

\section{Temporal Reference Resolution Rules}

\subsection{Introduction}

This document provides a detailed algorithm for temporal reference
resolution as used in the Artwork system developed at New Mexico State
Univsersity. The remainder of this section covers conventions used in
the algorithm. The algorithm itself is in Section \ref{rules}.

\subsubsection{Abbreviations}
\begin{enumerate}
\item {\bf SpanUtt}  = Spanish Utterance; a list of the Spanish words 
                  occurring in a sentence. For instance: 
                  ``cinco de marzo no'' would be represented as 
                  [cinco,de,marzo,no].
 
\item {\bf TU}  = Temporal Unit.
\item {\bf ILT} = A structured representation of the meaning of an
{\bf SpanUtt}, possibly containing a TU.
The ILT representation is defined by the Enthusiast project 
(Levin et al. 1995).

\item {\bf DE}  = Discourse Entity; Composed of 
a {\bf SpanUtt} and an {\bf ILT}.
 
\item {\bf FL}  = Focus List (ordered list of {\bf DE}'s).
 
\item {\bf RF}  = Reference Frame; a Temporal Unit used as a temporal frame of
                  reference.  
 
\item {\bf CF}  = Certainty Factor.  

\item {\bf timeValue} = One of \{{\bf day, week, weekend, month, year}\}.

\end{enumerate}

\subsubsection{Functions}

\begin{enumerate}
\item {\bf retrieveField}($FieldName$, $Structure$): returns the filler for 
                                   {\em FieldName} in {\em Structure}, if 
                                   {\em Structure} contains a $FieldName$, 
                                   otherwise NULL.

\item {\bf retrieveUtterance}($DE$): returns the Spanish utterance 
                                    from {\em DE}, if one 
                                    exists, otherwise NULL.


\item {\bf dateCopy}($TU$):   Copies the start dates of a $TU$ to their 
                              corresponding end dates.  The function
                              returns the modified $TU$ and a $CF$, as a
                              tuple.  The $CF$ is based on the combination
                              of fields copied.  

\item {\bf distanceFactor}($DE$):  Returns a number reflecting how far
                                   back on the focus list $DE$ is.

\item {\bf futureRange}($N\_ILT$): Returns a range in the 
                                   future consistent with the ILT.  
                                   For instance, ``second week in August'',
                                   interpreted as the second Monday
                                   through the following Friday
                                   in August, closest in the future.  If
                                   successful, a new TU is returned,
                                   otherwise, NULL.

\item {\bf last}($timeValue$, $RF$):  returns the last {\em timeValue}
                                      before reference frame {\em RF}.  
\begin{quote}
\begin{tabular}{lll} 
Ex: & last({\bf week}, [Tues 20 Aug 1996] \\
    & Mon-Fri, 12-16th, August, 1996 \\
\end{tabular}
\end{quote}

\item {\bf lastFLRange}($timeValue$, $N$): $N$ is either '1', or '2'.
        If $N$ $=$ '1' then {\bf lastFLRange} returns the $TU$ on the
        focus list
        that most closely 
matches {\em timeValue} from the focus list.  If $N$ $=$ '2', one
matching $TU$ on the list is skipped.

\item {\bf merge}($TU_1$, $TU_2$): if there exist 
                       conflicting fields in $TU_1$ and $TU_2$, return NULL.
                       Otherwise return a $TU$ which contains the union of all 
                       fields of $TU_1$ and $TU_2$.  

\item {\bf mergeUpper}($TU_1$, $TU_2$): the same as {\bf merge}, 
                            except only fields of the same or 
                            less specific levels than the most specific field 
                            in $TU_1$ are considered. 

\item {\bf mostSpecific}($X$, $TU$): $X$ is either {\bf start}, {\bf end},
                                   or {\bf both} to indicate the
                                   starting time, end time, or both
                                   the starting and end time,
                                   respectively.  The function returns
                                   the specificity level of the most
                                   specific field of $TU$, where {\em
                                   month} is level 1.  Ex:
                                   mostSpecific(start, $TU$) returns
                                   the specificity level of the most
                                   specific field in the start time of
                                   $TU$, and mostSpecific($both$,
                                   $TU$) returns the specificity
                                   level of the most specific field of
                                   the entire $TU$.

\item {\bf next}($timeValue$, $RF$):  returns the next {\em timeValue} that 
                                      follows the reference frame {\em RF}.

\item {\bf nextInclToday}($timeValue$, $RF$):  Same as {\bf next}, 
                                        but this version considers 
                                        ``today'' as a possibility.

\item {\bf resolveDeictic}($TU$, $todaysDate$): resolves the deictic term $TU$
                                                with respect to the dialog
                                                date.
\item {\bf applyRule}($TU$, $RuleName$ [, $subcase$]):  Invokes rule
{\em RuleName}.  Returns the return value of
{\em RuleName}.  

\item {\bf this}($timeValue$): Returns the current $timeValue$ with respect 
                               to the conversation date.
\begin{quote}
\begin{tabular}{lll} 
Ex: & Dialog date is Thursday, 22th, August, 1996 \\
    & ``This week has been long.'', \\
    & Returns Mon-Fri, 19th-23rd, August, 1996 \\
\end{tabular}
\end{quote}

\item {\bf isDeictic}($TU$):  returns TRUE if $TU$ contains
deictic information, otherwise NULL.

\item {\bf isRange}($X$, $TU$): returns TRUE, if $TU$ is an $X$, where $X$ = 
                            \{{\bf week, day, time\}}, otherwise  
                            returns NULL.

\item {\bf moreEquSpecific}($TU_1$, $TU_2$): returns TRUE if $TU_1$ is either 
                               more specific than $TU_2$, or if the two 
                               have the same level of specificity, 
                               otherwise, NULL.

%
\item {\bf moreSpecificLow}($TU_1$, $TU_2$): same as {\bf moreEqu Specific}, 
                                             but only tests levels of less
                                             or equal specificity as 
                                             time$\_$of$\_$day.

\end{enumerate}

\subsubsection{Typographic Conventions}
Variables are {\em italicized} (e.g., {\em TU, startTimes}).
Variable values (e.g., {\bf month, date}), keywords 
(e.g., {\bf if, then, else}), and conceptual notations
are in {\bf bold-face}.
Function names are in roman (e.g., merge({\em TU$_1$}, {\em TU$_2$})).
{\bf X$\rightarrow$Y} refers to the field(s) Y within X 
(e.g., {\em TU$\rightarrow$startTimes, TU$\rightarrow$endTimes, TU$\rightarrow$name})
\% Denotes comments in rules.
{\bf forward-looking-adjective} are adjectives that indicate a time
in the future.  For instance, ``next'', ``following'', or in Spanish
``proximo'', ``siguiente'', ``viene''.

\subsection{Rules}
\label{rules}
Rules are invoked in two different ways: one way for rules that do not
access the focus list, and one for those that do.  For rules that do
not access the focus list the procedure is as follows.
\begin{enumerate}
\item Extract all of the $TU$s in the $DE$.
\item Apply the rule to each $TU$.
\item Put the resulting $TU$s back in the same order as they appeared in 
the original $TUL$.
\end{enumerate}
For rules the do access the focus list, a similar process is
performed, but with one important difference: When accessing the
specific rule for the first Temporal Unit, which entity off
the focus list that was used is recorded, and the remaining temporal
units are forced to use the same one.

\subsubsection{Rules for deictic relations}
\label{sec:ANA_Rules}
\newcounter{counterNA_Rel01} 
\begin{list}{\bf Deictic Rule \arabic{counterNA_Rel01}:} {\usecounter{counterNA_Rel01}}
\item  {\em (A type of relation {\bf D-simple})} \\ 
{\em A deictic expression is resolved into a time interpreted with respect to the dialog date (e.g., ``Tomorrow'', ``last week'').}
\begin{tabbing}
mm\=mm\=mm\=mm\=mm\=mm\=mm\=mm\= \kill 
{\bf if} isDeictic({\em TU}) {\bf then} \\
\>              {\bf return} $\langle$0.9, merge({\em TU}, resolveDeictic({\em TU},
todaysDate))$\rangle$ \\
{\bf else fail} \\
\% {\em Ex:  ``Let's meet tomorrow'', with an interpretation of Mon 24 Sept}\\
\end{tabbing}

\item {\em (A type of relation {\bf D-frame-of-reference})} \\ 
\newcounter{counterA_Rel01sub} 
{\em A forward time is calculated with respect to today as a frame of reference.}
\begin{list}{\bf Subcase \roman{counterA_Rel01sub}:} {\usecounter{counterA_Rel01sub}}
\item
\begin{tabbing}
mm\=mm\=mm\=mm\=mm\=mm\=mm\=mm\= \kill \\
{\bf if not} isDeictic({\em TU}) {\bf then} \\
\>              {\em SpanUtt} $=$ retrieveUtterance(ILT) \\
\>         {\bf if} {\bf forward-looking-adjective} $\cap$ {\em SpanUtt} $\neq$ \{\} {\bf then} \\
\>\>            {\bf if}(mostSpecific({\bf start}, $TU$) $\leq$ {\bf date}) {\bf then} \\
\>\>\>          new$\_${\em TU} $=$ next(mostSpecific(start,{\em TU}), todaysDate) \\
\>\>\>          {\bf return}$\langle$0.2, merge({\em TU}, {\em new$\_$TU})$\rangle$ \\
\>\>            {\bf else fail} \\
\>         {\bf else fail} \\
{\bf else fail} \\
\% {\em Ex:  ``How about if we meet next Monday?'', interpreted as Monday, Sep 30}
\end{tabbing}
\item
\begin{tabbing}
mm\=mm\=mm\=mm\=mm\=mm\=mm\=mm\= \kill \\
{\bf if not} isDeictic({\em TU}) {\bf then} \\
\>         {\bf if}(mostSpecific({\bf start}, $TU$) $\leq$ {\bf date}) {\bf then} \\
\>\>            {\em new$\_$TU} $=$ nextInclToday(mostSpecific(start,{\em TU}), todaysDate) \\
\>\>            {\bf return}$\langle$0.3, merge($TU$,{\em new$\_$TU})$\rangle$ \\
\>         {\bf else fail} \\
{\bf else fail} \\
\% {\em Ex:  ``Hmmm, how about Monday?'' interpreted as the {\em current} Monday,}
\\{\em that is Mon 23 Sept.  This rule differers from the previous one only}\\
{\em in that the current date is also taken into consideration.} \\
\end{tabbing}
\end{list}

\noindent
The following cover special cases not discussed in the body of the
paper.

\item {\em (A type of relation {\bf D-simple})}\\
\begin{tabbing}
mm\=mm\=mm\=mm\=mm\=mm\=mm\=mm\= \kill 
{\bf if not} isDeictic({\em TU}) {\bf then} \\
\>{\bf if}({\bf the$\_$rest$\_$of} $\in$ {\em TU$\rightarrow$Spec}) {\bf then} \\
\>\>            {\em newTU} $=$ mergeAll(the$\_$rest$\_$of({\em Name})\\
\>\>    \% {\em Ex: ``I can meet Tuesday'', resolved to Tue 25 Sept} \\
\>\>    {\em ``Good, I can meet for the rest of the week as well'',} \\
\>\>    {\em resolved to Wed 26 Sept - Fri 28 Sept''}\\  
{\bf else if}({\bf the$\_$end$\_$of} $\in$ {\em Spec}) $\wedge$ ({\bf this} $\in$ {\em Spec}) {\bf then} \\
\>\>            {\em newTU} $=$ the$\_$end$\_$of({\em Name}) \\
\>\>    \% {\em Ex: ``I can meet Monday'', resolved to Mon 24 Sept} \\
\>\>    {\em ``No -- better at the end of the week.'',} \\
\>\>    {\em resolved to Thu 27 Sept - Fri 28 Sept''}\\  
{\bf else if}({\bf the$\_$end$\_$of} $\in$ {\em Spec}) $\wedge$ ({\bf next} $\in$ {\em Spec}) {\bf then} \\
\>\>            {\em newTU} $=$ the$\_$end$\_$of$\_$next({\em Name}) \\
\>\>    \% {\em Ex: ``I can meet at the end of next month'', resolved to Mon 21 Oct - Thu 31 Oct}\\
\>\>    {\em (The last ten days of the coming month)} \\
{\bf else if}({\bf last} $\in$ {\em Spec}) {\bf then} \\
\>\>            {\em newTU} $=$ last({\em Name}, {\em TU}) \\
\>\>    \% {\em Ex: ``Last Thursday's meeting was very productive.'', resolved to Thu 17 Sept}\\
{\bf else if}({\bf this} $\in$ {\em Spec}) $\wedge$ ({\bf coming} $\in$ {\em Spec}) {\bf then} \\
\>\>            {\em newTU} $=$ next({\em Name}, {\em TU}) \\
\>\>    \% {\em Ex: ``This coming Thursday is good.'', resolved to Thu 27 Sept}\\
{\bf else if}({\bf this} $\in$ {\em Spec}) {\bf then} \\
\>\>            {\em newTU} $=$ this({\em Name}, {\em TU}) \\
\>\>    \% {\em Ex: ``This has been a good month.'', resolved to Sat 1 Sept - Sun 30 Sept}\\
{\bf else} \\
\>\>    {\em newTU} $=$ futureRange($TU$) \\
\>\>    {\bf if}($newTU$) = {\bf null then fail}\\
\>\>    \% {\em Ex:  ``I can meet second week in October'',}\\
\>\>    {\em under the interpretation Mon 8 Oct to Fri 12 Oct.} \\
{\bf return} $\langle$0.5, merge($TU$, {\em newTU})$\rangle$ \\
\end{tabbing}

\item {\em (A special case for handling end times)}
{\em The mentioned time is not an interval; resulting frame's 
end dates correspond to the starting dates.} 
\begin{tabbing}
mm\=mm\=mm\=mm\=mm\=mm\=mm\=mm\= \kill 
{\bf if} (simpleTime({\em TU}) $\wedge$ \{{\bf sday, sdate, smonth}\} is not {\bf null}) {\bf then}\\ 
\>      $\langle${\em newTU, CF}$\rangle$ $=$ dateCopy({\em TU}) \\
\>      {\bf return} $\langle${\em CF}, merge($TU$, {\em newTU})$\rangle$ \\
{\bf else fail} \\
\% {\em Ex:  ``Let's meet Friday the 27 of September'', under the interpretation that the} \\
{\em start time and the end time both refer to Fri 27 Sept.} \\
\end{tabbing}

\item {\em (A type of relation {\bf D-simple})}\\
\begin{tabbing}
mm\=mm\=mm\=mm\=mm\=mm\=mm\=mm\= \kill 
{\bf if} at least one of \{{\em TU$\rightarrow$SMIN, TU$\rightarrow$SHOUR, TU$\rightarrow$STIME\_DAY}\} $\neq$ {\bf null} $\wedge$\\
current utterance does not contain a date {\bf then} \\
\>              \% Make sure that no $FL$ entry refers to {\bf time$\_$of$\_$day} or lower \\
\>              {\bf if} ($\forall$ TU $\in$ $FL$ : mostSpecific({\bf both}, $TU_{fl}$) $\geq$ {\bf time$\_$of$\_$day}) {\bf then} \\
\>\>            $today\_TU$ = today's date\\
\>\>            {\bf return} $\langle$0.5, merge($TU$, $today\_TU$)$\rangle$ \\
\>                {\bf else fail}\\
{\bf else fail} \\
\begin{tabular}{ll}
\% {\em Ex:} & {\em ``How about 3:00?'' resolved to 3:00 Mon 23 Sept}
\end{tabular}
\end{tabbing}

\item {\em (A type of relation {\bf D-simple})}\\
\begin{tabbing}
mm\=mm\=mm\=mm\=mm\=mm\=mm\=mm\= \kill 
{\bf if}({\em TU$\rightarrow$Name} $=$ \{{\bf weekend}\}) $\wedge$ ({\em TU$\rightarrow$Spec} $\in$ \{{\bf next, this, last}\}) {\bf then} \\
\>              if({\em TU$\rightarrow$Spec} $=$ {\bf next} {\bf then} \\
\>\>                    $new\_TU$ = next({\bf weekend}) \\
\>              if({\em TU$\rightarrow$Spec} $=$ {\bf this} {\bf then} \\
\>\>                    $new\_TU$ = this({\bf weekend}) \\
\>              if({\em TU$\rightarrow$Spec} $=$ {\bf last} {\bf then} \\
\>\>                    $new\_TU$ = last({\bf weekend}) \\
\>      {\bf return} $\langle$0.8, merge($TU$, $new\_TU$)$\rangle$ \\
\>                {\bf else fail}\\

{\bf else fail}\\
\end{tabbing} 
\begin{tabular}{ll}
\% {\em Ex:} & {\em ``Is next weekend OK?'' resolved to Sat 29-Sun 30 Sept}
\end{tabular}
\end{list}

\subsubsection{Rules for anaphoric relations}
\label{sec:AA_Rules}
\newcounter{counterA_Rel01} 
\begin{list}{\bf Anaphoric Rule \arabic{counterA_Rel01}:} {\usecounter{counterA_Rel01}}
\item {\em (A type of relation {\bf A-co-reference})}  \\ 
{\em The times discussed are similar; the resulting frame is the union of both times.}
\begin{tabbing}
mm\=mm\=mm\=mm\=mm\=mm\=mm\=mm\= \kill 
for each non-empty Temporal Unit $TU_{fl}$ from $FL$, starting with most recent\\
\>        {\bf if} (moreEquSpecific($TU$, $TU_{fl}$)) $\wedge$ (merge($TU$, $TU_{fl}$) $\neq$ {\bf null}) {\bf then}\\
\>\>              {\bf return} $\langle$0.8 - distanceFactor($TU_{fl}$), 
merge({\em TU}, $TU_{fl}$)$\rangle$ \\
\>        {\bf else fail} \\
\begin{tabular}{ll} 
\% {\em Ex:} & {\em``Let's meet Tuesday.'' resolved to Tue 24 Sept} \\
          & {\em ``How about 2?'' resolved to 2pm Tue 24 Sept } \\
\end{tabular}
\end{tabbing}
\end{list}

\addtocounter{counterA_Rel01}{1}
\begin{list}{\bf Anaphoric Rule \arabic{counterA_Rel01}:} {}
\item {\em (A type of relation {\bf A-less-specific})}  \\ 
{\em The current utterance evokes a time that includes the time evoked by a previous time, and the current time is less specific.}
\begin{tabbing}
mm\=mm\=mm\=mm\=mm\=mm\=mm\=mm\= \kill 
for each non-empty Temporal Unit $TU_{fl}$ from $FL$, starting with most recent\\
\>     {\bf if not} (moreEquSpecific({\em TU}, $TU_{fl}$)) $\wedge$ (moreEquSpecific($TU_{fl}$, {\em TU})) {\bf then} \\
\>\>        {\bf if} (merge($TU$, $TU_{fl}$) $\neq$ {\bf null}) {\bf then} \\
\>\>\>            {\bf return} $\langle$0.8 - distanceFactor({\em
TU}), 
merge({\em TU}, $TU_{fl}$)$\rangle$ \\
\>\>              {\bf else fail} \\
\>        {\bf else fail} \\
\begin{tabular}{ll} 
\% {\em Ex:} & {\em``Let's meet Tuesday.'' resolved to Tue 24 Sept} \\
          & {\em ``How about 2?'' resolved to 2pm Tue 24 Sept } \\
          & {\em ``On Tuesday?'' resolved to Tue 24 Sept } \\
          & {\em  This subcase differs from the previous subcase in that} \\
          & {\em  in this case a relationship can be made to hold if the} \\
          & {\em current utterance is less specific than the one}\\
          & {\em on the focus list.} \\
\end{tabular}
\end{tabbing}
\end{list}

\addtocounter{counterA_Rel01}{1}
\begin{list}{\bf Anaphoric Rule \arabic{counterA_Rel01}:} {}
\item {\em (A type of relation {\bf A-frame-of-reference})}  \\ 
{\em A forward time is calculated with respect to a time on the Focus List as a frame of reference.}
\newcounter{counterA_Rel03sub} 
\begin{list}{\bf Subcase \roman{counterA_Rel03sub}:} {\usecounter{counterA_Rel03sub}}
\item
\begin{tabbing}
mm\=mm\=mm\=mm\=mm\=mm\=mm\=mm\= \kill \\
{\bf if not} isDeictic({\em TU}) {\bf then} \\
\>      {\bf if} \{{\bf forward-looking-adjective}\} $\cap$ {\em ILT$\rightarrow$SpanUtt} $\neq$ \{\} {\bf then} \\
\>\>    {\bf if}(leastSpecific({\em TU$\rightarrow$startFields}) $\leq$ {\bf date}) {\bf then} \\
\>\>\>          for each non-empty Temporal Unit $TU_{fl}$ from $FL$, starting with most recent \\
\>\>\>\>                          {\bf if} ({\em TU$\rightarrow$name} $=$ \{{\bf week}\} {\bf then} \\
\>\>\>\>\>{\em Base$\_$CF} $=$ 0.4 \\
\>\>\>\>                {\bf else} \\
\>\>\>\>\>                      {\em Base$\_$CF} $=$ 0.3 \\
\>\>\>\>                          {\bf if} moreEquSpecific({\em TU}, $TU_{fl}$) {\bf then} \\
\>\>\>\>\>                      {\em RF} $=$ retrieveStartDate({\em $TU_{fl}$}) \\
\>\>\>\>\>                      {\em new$\_$TU} $=$ next(mostSpecific(start,$TU$), {\em RF}) \\
\>\>\>\>\>                      {\bf return} $\langle${\em Base$\_$CF} - distanceFactor({\em TU}), merge($TU$, $new\_TU$)$\rangle$ \\
\>\>\>\>                          {\bf else fail} \\
\>\>              {\bf else fail} \\
\>        {\bf else fail} \\
{\bf else fail} \\
\begin{tabular}{ll} 
\% {\em Ex:} & {\em``Let's meet next Monday.'' resolved to Mon 30 Sept} \\
          & {\em ``Tuesday is better for me.'' resolved to Tue 1 Oct} \\
\end{tabular}
\end{tabbing}
\item
\begin{tabbing}
mm\=mm\=mm\=mm\=mm\=mm\=mm\=mm\= \kill \\
{\bf if not} isDeictic({\em TU}) {\bf then} \\
\>      {\bf if}(leastSpecific({\em TU$\rightarrow$startFields}) $\leq$ {\bf date}) {\bf then} \\
\>\>            for each non-empty Temporal Unit $TU_{fl}$ from $FL$, starting with most recent \\
\>\>\>                    {\bf if} moreEquSpecific({\em TU}, $TU_{fl}$) {\bf then} \\
\>\>\>\>                        {\em RF} $=$ retrieveStartDate($TU_{fl}$) \\
\>\>\>\>                        {\em new$\_$TU} $=$ nextInclToday(mostSpecific(start,$TU$), $RF$) \\
\>\>\>\>                        {\bf return} $\langle$0.35 -
distanceFactor({\em TU}), merge($TU$, $new\_TU$)$\rangle$ \\ 
\>\>\>                    {\bf else fail} \\
\>        {\bf else fail} \\
{\bf else fail} \\
\begin{tabular}{ll} 
\%& {\em This is nearly identical to the previous case, but we're not} \\
          & {\em looking for a {\bf forward-looking-adjective}, and} \\
          & {\em today's date is considered.} \\
{\em Ex:} & {\em``Let's meet next Monday.'' resolved to Mon 30 Sept} \\
          & {\em ``Monday is good.'' resolved to Mon 30 Sept} \\
          & {\em Since the second sentence doesn't provide clues} \\
          & {\em about the Monday (such as ``next'', ``the following'' etc.),} \\
          & {\em the current Monday on the focus list is used as opposed} \\
          & {\em to the Monday following Monday the 30th.} \\
\end{tabular}
\end{tabbing}
\end{list}

\addtocounter{counterA_Rel01}{1}
\begin{list}{\bf Anaphoric Rule \arabic{counterA_Rel01}:} {} 
\item {\em (A type of relation {\bf A-modify})} \\ 
{\em The current time is a modification of a previous time}
\begin{tabbing}
mm\=mm\=mm\=mm\=mm\=mm\=mm\=mm\= \kill 
{\bf if not} isDeictic({\em TU}) {\bf then} \\
\>              {\bf if}(mostSpecific({\bf both}, {\em TU}) $\geq$ {\bf time$\_$of$\_$day}) \\
\>\>                    for each non-empty Temporal Unit $TU_{fl}$ from $FL$, starting with most recent \\
\>\>\>                    {\bf if} (moreSpecificLow({\em TU}, {\em $TU_{fl}$}) {\bf then} \\
\>\>\>\>                        {\bf return} $\langle$0.35 -
distanceFactor({\em TU}), mergeUpper({\em $TU_{fl}$}, {\em
TU})$\rangle$ \\
\>\>\>            {\bf else fail} \\
\>        {\bf else fail} \\
{\bf else fail} \\
\begin{tabular}{ll}
\% {\em Ex:} & {\em``Let's meet Tuesday at 2:00.'' resolved to 2 pm, Tue 24 Sept} \\
          & {\em ``3:00 is better for me.'' resolved to 3 pm, Tue 24 Sept} \\
\end{tabular}
\end{tabbing}
\end{list}

\addtocounter{counterA_Rel01}{1}

\noindent
The following anaphoric rules do not appear in the paper.

\begin{list}{\bf Anaphoric Rule \arabic{counterA_Rel01}:} {} 
\item {\em (A type of relation {\bf A-frame-of-reference})}
\begin{tabbing}
mm\=mm\=mm\=mm\=mm\=mm\=mm\=mm\= \kill 
{\bf if} ({\em TU$\rightarrow$startFields} $=$ \{\}) $\wedge$ ({\em TU$\rightarrow$endFields} $\neq$ \{\}) {\bf then} \\
\>              for each non-empty Temporal Unit $TU_{fl}$ from $FL$, starting with most recent\\
\>\>            {\bf if} moreEquSpecific({\em TU}, $TU_{fl}$) {\bf then} \\
\>\>\>                  {\em new$\_$TU} $=$ mergeUpper({\em $TU_{fl}$}, {\em TU}) \\
\>\>\>                  {\bf if}(leastSpecific({\bf end}, $TU$) $\leq$ {\bf date}) {\bf then} \\
\>\>\>\>                        {\em RF} $=$ retrieveStartDate({\em $TU_{fl}$}) \\
\>\>\>\>                        {\em new$\_$TU}$\rightarrow$endFields $=$ next(mostSpecific({\em endFields, $TU$}), {\em RF}) \\
\>\>\>\>                        {\bf return} $\langle$0.5 - distanceFactor($TU$),$new\_TU$)$\rangle$  \\
\>\>\>            {\bf else fail} \\
\>\>      {\bf else fail} \\
{\bf else fail} \\
\begin{tabular}{ll} 
\% {\em Ex:} & {\em``How about Tuesday the 25th at 2?'' resolved to 2 pm, Tue 25 Sept} \\
          & {\em ``I am busy until Friday'', resolved to Tue 25-Friday 28 Sept} \\
\end{tabular}
\end{tabbing}

\addtocounter{counterA_Rel01}{1}
\item {\em (A type of relation {\bf A-frame-of-reference})}\\
\newcounter{counterA_Rel10sub} 
\begin{list}{\bf Subcase \roman{counterA_Rel10sub}:} {\usecounter{counterA_Rel10sub}}
\item
\begin{tabbing}
mm\=mm\=mm\=mm\=mm\=mm\=mm\=mm\= \kill \\
{\bf if not}(isDeictic({\em TU})) {\bf then} \\
\>{\bf if}({\em TU$\rightarrow$Spec} $\in$ \{{\bf that, same, all$\_$range, less$\_$than, more$\_$than, long}\}) \\
\>      $\wedge$ ({\em TU$\rightarrow$Name} $\in$ \{{\bf month, week, day, time}\}) {\bf then} \\
\>\>              {\em $TU_{fl}$} $=$ lastFLRange({\em Name}) \\
\>\>              {\bf if} (mergeUpper({\em $TU_{fl}$}, {\em TU})) $\neq$ {\bf null} {\bf then} \\
\>\>\>                  {\bf return}$\langle$0.5,mergeUpper({\em $TU_{fl}$}, {\em TU})$\rangle$ \\
\>\>              {\bf else if} (merge({\em $TU_{fl}$}, {\em TU}))$\neq$ {\bf null} {\bf then} \\
\>\>\>                  {\bf return}$\langle$0.5,merge({\em $TU_{fl}$}, {\em TU})$\rangle$ \\
\>\>              {\bf else fail}\\
\>        {\bf else fail}\\
{\bf else fail} \\
\begin{tabular}{ll} 
\% {\em Ex:} & {\em``Let's meet Oct 8th to Oct 11th'' resolved to Tue 8 Oct - Fri 11 Oct} \\
          & {\em ``that week sounds good'' resolved to Tue 8 Oct - Fri 11 Oct} \\
\end{tabular}
\end{tabbing}
\item
\begin{tabbing}
mm\=mm\=mm\=mm\=mm\=mm\=mm\=mm\= \kill \\
{\bf if not}(isDeictic({\em TU})) {\bf then} \\
\>      {\em Name} $=$ retrieveField(name, {\em TU}) \\
\>      {\em Spec} $=$ retrieveField(specifier, {\em TU}) \\
\>      {\bf if}({\bf the$\_$rest$\_$of} $\in$ {\em Spec}) $\wedge$ ({\bf that} $\in$ {\em Spec}) {\bf then} \\
\>\>            {\em startTU$\rightarrow$startFields} $=$ applyRule($TU$, ANA3, 'i') \\
\>\>            {\em endTU$\rightarrow$endFields} $=$ next({\em Name}, {\em startFields}) \\
\>\>            {\em newTU} $=$ merge({\em startTU, endTU}) \\
\>      {\bf else if}({\bf the$\_$end$\_$of} $\in$ {\em Spec}) $\wedge$ ({\bf that} $\in$ {\em Spec}) {\bf then} \\
\>\>            {\em RF} $=$ applyRule($TU$, ANA3, 'i') \\
\>\>            {\em newTU} $=$ the$\_$end$\_$of({\em RF}) \\
\>\>            {\bf return} $\langle$0.5,{\em newTU}$\rangle$ \\
\>      {\bf else fail}\\
{\bf else fail} \\
\begin{tabular}{ll}
\% {\em Ex:} & {\em``Let's meet Oct 8th to Oct 11th'' resolved to Tue 8 Oct - Fri 11 Oct} \\
          & {\em ``The end of that week is better'' resolved to Thu 10 Oct - Fri 11 Oct} \\
\end{tabular}
\end{tabbing}
\end{list}

\addtocounter{counterA_Rel01}{1}
\item {\em (A type of relation {\bf A-co-reference})}\\
\begin{tabbing}
mm\=mm\=mm\=mm\=mm\=mm\=mm\=mm\= \kill 
{\bf if}({\bf other} $\in$ {\em TU$\rightarrow$Spec}) $\wedge$ 
({\bf indefinite} not in {\em TU$\rightarrow$Spec}) {\bf then} \\
\>      {\bf if}({\em Name} $\in$ \{{\bf month, week, day, time}\}) {\bf then}\\
\>\>            {\em newTU} $=$ lastFLRange($TU$,2) \\
\>\>            if(merge($newTU$, $TU$) $\neq$ {\bf null}) {\bf then} \\
\>\>\>                  {\bf return} $\langle$0.7 - 
distanceFactor({\em newTU}),merge($newTU$,$TU$)$\rangle$ \\
\>                {\bf else fail}\\
{\bf else fail} \\
\begin{tabular}{ll}
\% {\em Ex:} & {\em ``Let's meet Tuesday.'' resolved to Tue 24 Sept} \\
          & {\em ``How about Thursday?'' resolved to Thu 26 Sept} \\
          & {\em ``No, the other day sounds better'' resolved to Tue 24 Sept}\\
\end{tabular}
\end{tabbing}

distanceFactor($TU_{fl}$), $new\_TU$$\rangle$ \\

\addtocounter{counterA_Rel01}{1}
\item {\em (A type of relation {\bf A-co-reference})}
\begin{tabbing}
mm\=mm\=mm\=mm\=mm\=mm\=mm\=mm\= \kill 
$Name:  $ $=$ retrieveField({\bf name}, {\em TU}) \\
$Spec$ $=$ retrieveField({\bf specifier}, {\em TU}) \\
{\bf if}($Name:  $ $\in$ \{{\bf week, day, time}\}) $\wedge$ ($Spec$ $\in$ \{{\bf that, plural, both$\_$of}\}) {\bf then} \\
\>          for each non-empty Temporal Unit $TU_{1-fl}$ from $FL$, starting with most recent \\
\>\>          for each non-empty Temporal Unit $TU_{2-fl}$ from $FL$, starting with $TU_{1-fl}$ \\
\>\>          {\bf if}(isRange($Name:  $, $TU_{1-fl}$)) $\wedge$ (isRange($Name:  $, $TU_{2-fl}$)) {\bf then}\\
\>\>\>      $Speaker1$ = retrieveField({\bf speaker}, SpanUtt) \\
\>\>\>      $Speaker2$ = retrieveField({\bf speaker}, SpanUtt) \\
\>\>\>      {\bf if}($Speaker1 = Speaker2$) {\bf then} \\
\>\>\>\>        {\bf if}($Name = $ {\bf day}) $\wedge$ ($TU_{1-fl} \wedge TU_{2-fl}$ are within the same week) $\vee$ \\
\>\>\>\>        {\bf if}($Name = $ {\bf week}) $\wedge$ ($TU_{1-fl} \wedge TU_{2-fl}$ are within the same year) $\vee$ \\
\>\>\>\>        {\bf if}($Name = $ {\bf day}) $\wedge$ ($TU_{1-fl} \wedge TU_{2-fl}$ are in the same day) {\bf then} \\
\>\>\>\>\>      {\bf return} $\langle$0.7, [merge($TU$, $TU_{1-fl}$), merge($TU$, $TU_{2-fl}$)]$\rangle$ \\
\>\>\>\>      {\bf else fail}\\
\>\>\>       {\bf else fail}\\
\>\>      {\bf else fail}\\
{\bf else fail} \\
\begin{tabular}{ll}
{\em Ex:} & {\em ``How about Tuesday and Wednesday?'' resolved to Tue 24 Sept, Wed 25 Sept}\\
          & {\em ``Those days sound fine.'' resolved to Tue 24 Sept, Wed 25 Sept}\\
\end{tabular}
\\\% This rule heuristically uses the last {\em two} occurrences of $Name:  $.
\end{tabbing}
\end{list}
\end{list}


\newpage
\section{Normalized form of a {\bf Temporal Unit}}

This appendix gives the specific structure, and possible information
contained within a normalized {\em Temporal Unit} (TU), as used in the
{\em Artwork} project, developed at New Mexico State University in
conjunction with the Computing Research Laboratory.  A list of
Temporal Units {\bf LTU} is a structure containing one or more TU's.

\subsection{Notation Used}
The following BNF-style convention is used:
\begin{enumerate}
\item {\bf atom} An {\bf atom} is a single entry in a list.  An atom cannot
be expanded any further.
\item {\underline {Non-terminal}} A {\underline {Non-terminal}} indicates a 
structure that is constructed by either one or more atoms, or of other non-terminals.
\item {\bf[...]} open-close angeled brackets denote the beginning
and end of a list.  For example, {[{\bf this, is, a, list}]} is a list
containing four {\bf atom}s.
\item * An asterisk denotes a list that may be repeated zero or more
times.  For example: [Repeat, this, list]* $\rightarrow$ [ ] (empty list) or
[{\bf repeat, this, list}],\dots, [{\bf repeat, this, list}].
\item + A plus sign denotes a list that occurs one or more times.
time.
\item {\bf $\mid$} The ``$\mid$'' sign indicates a choice.  For example:
{\underline {Person}} = {\bf man} $\mid$ {\bf woman}
\end{enumerate}

\subsection{High Level Format of a When Frame}
All structures in the Artwork project, adhere with the following
format:
\begin{quote}
{\center [{\underline {Structure Name}}, {\underline {Structure Filler}}]}\\
\end{quote}

\noindent
For example,

\begin{quote}
\noindent{[{\bf Structure Name}, {\underline {Structure Filler}}]}\\
\noindent{[{\bf sday$\_$week}, {\underline {SDAY$\_$WEEK}}]}\\
\noindent{[{\bf sday$\_$week}, {\bf thursday}]}.\\
\end{quote}

\noindent
The general format of an {\underline {LTU}} follows:

\begin{quote}
{\center {\underline {LTU}} $=$ [{\bf when-frame}, {\underline {TU}}*] \\}
\end{quote}

\noindent
Thus, the name of the structure is {\em when-frame} and the filler is
either zero ([ ]), or more TU.

\subsection{Detailed Format of a TU}

\begin{figure}[H]
\begin{center}

\begin{tabbing}
mm\=mm\=mm\=mm\=mm\=mm\=mm\=mm\= \kill \\
\>		  [{\bf when}, \\
\>\>                    [{\bf connective}, {\underline {CONNECTIVE}}], \\
\>\>                    [{\bf gen$\_$spec}, {\underline {GEN$\_$SPEC}}], \\
\>\>                    [{\bf duration}, [length, {\underline {DURATION}}], [dur$\_$specifier, {\underline {DUR$\_$SPECIFIER}}], \\
\>\>                    [{\bf name}, {\underline {NAME}}], \\
\>\>                    [{\bf interval}, \\ 
\>\>\>                   [{\bf specifier}, [{\underline {SPECIFIER}}]], \\
\>\>\>                   [{\bf start}, \\
\>\>\>\>                  [{\bf sday$\_$status}, [{\underline {DAY$\_$VALUE}}, {\underline {DAY$\_$ORIGIN}}], \\
\>\>\>\>\>                 [{\bf sday$\_$week}, {\underline {DAY$\_$WEEK}}], \\
\>\>\>\>\>                 [{\bf sday}, {\underline {DAY}}], \\
\>\>\>\>\>                 [{\bf stime$\_$day}, {\underline {TIME$\_$DAY}}], \\
\>\>\>\>\>                 [{\bf sam$\_$pm}, {\underline {AM$\_$PM}}]], \\
\>\>\>\>                  [{\bf smonth}, [{\underline {MONTH$\_$VALUE}}, {\underline {MONTH$\_$ORIGIN}}]], \\
\>\>\>\>                  [{\bf shour$\_$status}, [{\underline {HOUR$\_$VALUE}}, {\underline {HOUR$\_$ORIGIN}}],\\ 
\>\>\>\>\>                 [{\bf stime$\_$adv}, {\underline {TIME$\_$ADV}}], \\
\>\>\>\>\>                 [{\bf shour}, {\underline {HOUR}}], \\
\>\>\>\>\>                 [{\bf smin}, {\underline {MIN}}]]], \\
\>\>\>                   [{\bf end}, \\
\>\>\>\>                  [{\bf eday$\_$status}, [{\underline {DAY$\_$VALUE}}, {\underline {DAY$\_$ORIGIN}}], \\
\>\>\>\>\>                 [{\bf eday$\_$week}, {\underline {DAY$\_$WEEK}}], \\
\>\>\>\>\>                 [{\bf eday}, {\underline {DAY}}], \\
\>\>\>\>\>                 [{\bf etime$\_$day}, {\underline {TIME$\_$DAY}}], \\
\>\>\>\>\>                 [{\bf eam$\_$pm}, {\underline {AM$\_$PM}}]], \\
\>\>\>\>                  [{\bf emonth}, [{\underline {MONTH$\_$VALUE}}, {\underline {MONTH$\_$ORIGIN}}]], \\
\>\>\>\>                  [{\bf ehour$\_$status}, [{\underline {HOUR$\_$VALUE}}, {\underline {HOUR$\_$ORIGIN}}], \\ 
\>\>\>\>\>                 [{\bf etime$\_$adv}, {\underline {TIME$\_$ADV}}], \\
\>\>\>\>\>                 [{\bf ehour}, {\underline {HOUR}}], \\
\>\>\>\>\>                 [{\bf emin}, {\underline {MIN}}]]]], \\
\>\>                   [{\bf modifiers}, {\underline {MODIFIER}}]] \\

\end{tabbing}

\end{center}
\caption{Structural layout of a Temporal Unit.}
\label{TU}
\end{figure}

Figure \ref{TU} illustrates the structural layout of a single TU.
The following tables and figures presents the possible values each
temporal field can take (Table \ref{general-fields}, Figure
\ref{general-values} and Table \ref{temporal-fields}).  The NAME field
indicates the name referred to in the fields in the above structured
TU.  The VALUE column illustrates the set of values that each entry
might take.  A {\bf null} value indicates that no information is
available.  The third column provides a description of the {\em name}
field.  An in-depth description about each field is given at the end
of the table.
The last field in the table, ``{\bf R}'' 
indicates whether the information can be repeated.  A ``{\bf Y}'' indicates 
that it can be, and an ``{\bf N}'' indicates that it cannot be.  If a field is 
repeated, it is preceded by the keyword ``{\bf *multiple*}'', otherwise it is 
not preceded by the keyword.
For example, \\

\begin{quote}
\noindent
dur$\_$specifier = \{{\bf *multiple*, the$\_$end$\_$of, second}\} and \\
\noindent
{dur$\_$specifier = \{{\bf the$\_$end$\_$of}\}} \\
\end{quote}

Since dur$\_$specifier can be repeated, it is indicated by the keyword
{\bf *multiple*}, as shown in the first example.  However, since there
is only one filler in the second example, the keyword {\bf *multiple*}
is omitted.  Fields that can take on a large number of values or need
special attention are underlined in the second column of the table,
and expanded at the end of this section.

A Temporal Unit comprises three parts: information about the time in
general (e.g., duration, name, etc.), the start time, and the end
time.  Table \ref{general-fields} shows information on the general
fields, and Figure \ref{general-values} indicates the values these
fields can take.  Table \ref{temporal-fields} shows fields applicable
to the starting time of a TU.  The ending time fields are similar to
the fields in Table \ref{temporal-fields}, so no table is shown for
the ending times.

\begin{table}[H]
\begin{center}

\noindent
\begin{tabular}{|c|p{5.5cm}|p{5.5cm}|c|}
\hline 
{\bf NAME} & \hspace{0.5cm} {\bf POSSIBLE VALUES} & {\bf DESCRIPTION} \\
\hline
{\underline {CONNECTIVE}} & {\raggedleft \bf and $\mid$ because $\mid$ but $\mid$ for$\_$example $\mid$ 
			         if $\mid$ or $\mid$ so $\mid$ then $\mid$ therefore $\mid$ unless $\mid$ 
				 that$\_$is$\_$to$\_$say $\mid$ null}
			  & how multiple TUs are connected \\
\hline 
{\underline {GEN$\_$SPEC}} & {\raggedleft \bf generic $\mid$ specific $\mid$ null}
			   & the genericity of a time \\
\hline 	
{\underline {DURATION}}   & { \raggedleft 0..MAXINT (only whole numbers) 
			    $\mid$ {\bf epsilon $\mid$ 
			    undetermined $\mid$ null} }
			  & the difference (in hours) between the start and
			    end time \\
\hline 
{\underline {DUR$\_$SPECIFIER}} & 
{\raggedright \bf \mbox{determined }$\mid$\mbox{ part$\_$determined}\\$\mid$ not-complete \\ }
				& the frame from which the duration 
				  information was retrieved unless
				  calculated or null; repeatable \\
\hline 
{\underline {NAME}}       & {\raggedleft \bf {\underline {Indexical}} $\mid$ 
			    {\underline {Special Name}} $\mid$ 
			    {\underline {Unit}} $\mid$ 
			    null}
			  & special information about the current
			    TU not necessarily
			    determinable by other fields. \\
\hline 
{\underline {SPECIFIER}}  & {\raggedleft \bf {\underline {Specifier}} $\mid$
			    0..MAXINT (only whole numbers)
			    $\mid$ null}
			  & Info \\

\hline 
\end{tabular}
\\

\end{center}
\caption{General fields of a Temporal Unit}
\label{general-fields}
\end{table}

\begin{figure}[H]

\noindent 
{\underline {Specifier}} ::= 
\begin{tabbing}
mm\=mm\=mm\=mm\=mm\=mm\=mm\=mm\= \kill 
\> {\bf  all$\_$member $\mid$ all$\_$range $\mid$ also $\mid$}
   {\bf  another $\mid$ any $\mid$ anytime $\mid$ approximate $\mid$} \\
\> {\bf  at$\_$least $\mid$ behind $\mid$ both$\_$of $\mid$}
   {\bf  concrete $\mid$ couple $\mid$ definite $\mid$ early $\mid$} \\
\> {\bf  either$\_$of $\mid$ even $\mid$ exact $\mid$ except $\mid$}
   {\bf  few $\mid$ first $\mid$ following $\mid$ fourth $\mid$} \\
\> {\bf  in$\_$front$\_$of $\mid$ indefinite $\mid$ last $\mid$ late $\mid$}
   {\bf  less$\_$than $\mid$ long $\mid$ middle $\mid$ more$\_$than $\mid$} \\
\> {\bf  most$\_$member $\mid$ most$\_$range $\mid$ necessary $\mid$}
   {\bf  negative $\mid$ next $\mid$ only $\mid$ other $\mid$ particular $\mid$} \\
\> {\bf  perhaps $\mid$ plural $\mid$ preceding $\mid$ same $\mid$ second $\mid$}
   {\bf  second$\_$last $\mid$ short $\mid$ some $\mid$ sometime $\mid$ } \\
\> {\bf  that $\mid$ the$\_$end$\_$of $\mid$ the$\_$rest$\_$of $\mid$}
   {\bf  third $\mid$ this $\mid$ what $\mid$ which } \\
\end{tabbing}

\noindent 
{\underline {Indexical}} ::= 
\begin{tabbing}
mm\=mm\=mm\=mm\=mm\=mm\=mm\=mm\= \kill 
\> {\bf now $\mid$ today $\mid$ tomorrow $\mid$ date $\mid$}
   {\bf time $\mid$ then $\mid$ that $\mid$ there $\mid$} \\
\> {\bf here $\mid$ later $\mid$ when $\mid$ what} \\
\end{tabbing}

\noindent 
{\underline {Special Name}} ::= 
\begin{tabbing}
mm\=mm\=mm\=mm\=mm\=mm\=mm\=mm\= \kill 
\> {\bf independence$\_$day $\mid$ thanksgiving $\mid$} 
   {\bf thanksgiving$\_$day $\mid$ labor$\_$day} \\
\end{tabbing}

\noindent 
{\underline {Unit}} ::= 
\begin{tabbing}
mm\=mm\=mm\=mm\=mm\=mm\=mm\=mm\= \kill 
\> {\bf minute $\mid$ hour $\mid$ day $\mid$}
   {\bf week $\mid$ month $\mid$ year $\mid$} \\
\> {\bf decade $\mid$ century $\mid$ lunch$\_$time $\mid$}
   {\bf working$\_$day $\mid$ weekend $\mid$ weekday $\mid$} \\
\> {\bf early $\mid$ late $\mid$ sometime $\mid$}
   {\bf anytime} \\
\end{tabbing}

\caption{Possible values for the general fields}
\label{general-values}
\end{figure}


\begin{table}[H]
\begin{center}

\begin{tabular}{|c|p{5.0cm}|p{5.0cm}|c|}
\hline 
{\bf NAME} & \hspace{0.5cm} {\bf POSSIBLE VALUES} & {\bf DESCRIPTION} & {\bf R}\\
\hline

{\underline {SDAY$\_$VALUE}} & {\bf determined $\mid$ part$\_$determined $\mid$
				    undetermined $\mid$ same $\mid$ null}
			     & amount of information available to
			       determine the start date
			     & {\bf N} \\
\hline 
{\underline {SDAY$\_$ORIGIN}} & {\bf when $\mid$ clarified $\mid$ topic $\mid$ null} 
			      & the frame from which the start day 
				information was retrieved
			      & {\bf Y} \\
\hline 
{\underline {SDAY$\_$WEEK}} & {\bf monday,\ldots, sunday $\mid$ null}
			    & the start week-day
 	 		    & {\bf N} \\
\hline 
{\underline {SDAY}}         & {\bf 1,\ldots, 31 $\mid$ null} 
			    & the start date
 			    & {\bf N} \\
\hline 
{\underline {STIME$\_$DAY}} & {\bf afternoon $\mid$ evening $\mid$ mid$\_$afternoon $\mid$
			           mid$\_$morning $\mid$ midnight $\mid$ morning $\mid$
			           night $\mid$ noon $\mid$ null}
			    & the general start time of a 24-hour period
 	 	 	    & {\bf N} \\
\hline 
{\underline {SAM$\_$PM}}    & {\bf am $\mid$ pm $\mid$ null}
			    & the start time (morning $\mid$ afternoon)
			    & {\bf N} \\
\hline 
{\underline {SMONTH$\_$VALUE}} & {\bf 1,\ldots, 12 $\mid$ null} 
			       & the starting month in a numerical format.  
				 E.g. $\mid$ January = 1,\ldots, December = 12
			       & {\bf N} \\
\hline 
{\underline {SMONTH$\_$ORIGIN}} & {\bf when $\mid$ clarified $\mid$ topic $\mid$ null}
				& the frame from which the start month 
				  was retrieved
  	  	 	        & {\bf Y} \\
\hline 
{\underline {SHOUR$\_$VALUE}} & {\bf determined $\mid$ part$\_$determined $\mid$
				     undetermined $\mid$ same $\mid$ null}
			      & amount of information available to
			        determine the start time
 	 	 	      & {\bf N} \\
\hline 
{\underline {SHOUR$\_$ORIGIN}} & {\bf when $\mid$ clarified $\mid$ topic $\mid$ null} 
			       & the frame from which the start hour 
				 information was retrieved
		               & {\bf Y} \\			       
\hline 
{\underline {STIME$\_$ADV}} & {\bf after $\mid$ at $\mid$ before $\mid$ from $\mid$ 'til $\mid$ to $\mid$ until $\mid$ null} 
			    & the adverb associated with the start time
		            & {\bf N} \\
\hline 
{\underline {SHOUR}} & {\bf 1,\ldots, 12 $\mid$ null}
		     & This field represents the start hour
 		       of a time.  
		     & {\bf N} \\
\hline 
{\underline {SMIN}} & {\bf 0,\ldots, 59}
		    & the start minutes associated with the start hour
		    & {\bf N} \\
\hline 
\end{tabular}

\end{center}
\caption{Temporals fields of a Temporal Unit}
\label{temporal-fields}
\end{table}


\subsection{Examples}
This section provides examples, and a brief discussion about selected
field representations given in Table \ref{general-fields} and Table
ref{temporal-fields}. See Figure \ref{examples}.

\begin{figure}[H]

{\underline {CONNECTIVE:}}
\begin{quote}
\begin{tabular}{lll}
Ex: & ``I can meet Wednesday or Thursday'' \\
    & {\em This sentence will produce two TUs, both of 
	which have}\\
    & {\em {\underline{CONNECTIVE}} set to {\bf or}}\\
\end{tabular}
\end{quote}

{\underline {GEN$\_$SPEC}:}
\begin{quote}
\begin{tabular}{lll}
Ex: & ``How about Wednesday?'' \\
    & {\em Since ``Wednesday'' is any wednesday, {\underline {GEN$\_$SPEC}}
	is set to {\bf generic}} \\
Ex: & ``OK, that Wednesday sounds good'' \\
    & {\em In this example, the speaker is referring to a specific Wednesday,
	and,} \\
    & {\em thus, {\underline {GEN$\_$SPEC}} is set to {\bf specific}} \\
\end{tabular}
\end{quote}

{\underline {DURATION}:}
\begin{quote}
\begin{tabular}{lll}
Ex: & ``I can meet from 2:00 until 4:00'' \\
    & {\em {\underline {DURATION}} is set to {\bf 2} (hours)} \\
\end{tabular}
\end{quote}

{\underline {NAME}:}
\begin{quote}
\begin{tabular}{lll}
Ex: & ``I can meet next Thursday'' \\
    & {\em {\underline {NAME}} is set to {\bf next}}\\
    & {\em (It gives specific information about the time)} \\
\end{tabular}
\end{quote}

{\underline {SPECIFIER}:}
\begin{quote}
\begin{tabular}{lll}
Ex: & ``The second week in August is good for me'' \\
    & {\em In this example, {\underline {SPECIFIER}} is set to {\bf week}} \\
    & {\em ({\underline {NAME}} = {\bf second})} \\
\end{tabular}
\end{quote}

{\underline {SDAY$\_$VALUE}:}
\begin{quote}
\begin{tabular}{lll}
Ex: & ``I can meet Thursday'' \\
    & {\em {\underline {SDAY$\_$VALUE}} is set to {\bf part-determined}, 
	since the exact Thursday} \\
    & {\em cannot be determined.} \\
Ex: & ``I can meet Thursday the 11$^{th}$ of August'' \\
    & {\em {\underline {SDAY$\_$VALUE}} is set to {\bf determined}, since} \\
    & {\em we have enough information about the Thursday in question} \\
\end{tabular}
\end{quote}

{\underline {SHOUR$\_$VALUE}:}
\begin{quote}
\begin{tabular}{lll}
Ex: & ``How about 4 or so?'' \\
    & {\em {\underline {SHOUR$\_$VALUE}} is set to {\bf part-determined},}\\
    & {\em since we don't know the exact time.  (In this example, the
	minute} \\
    & {\em information is missing)} \\
Ex: & ``How about next Thursday'' \\
    & {\em {\underline {SHOUR$\_$VALUE}} is set to {\bf undetermined},
	since no} \\
    & {\em {information about the time is available.}}\\
\end{tabular}
\end{quote}

\caption{Examples of field values for temporal units}
\label{examples}
\end{figure}


\newpage
\section{Coverage of Temporal Expressions}

The BNF grammar in Figure \ref{BNF-coverage} describes the core set of
the temporal expressions handled by our temporal reference resolution
system. To simplify the grammar, some variations are not
specified. Combinations of the expressions are also allowed, provided
that this doesn't lead to contradictory temporal interpretations. This
is based on English transcription of the Spanish phrases that are
covered by the CMU semantic parser. Note that conjunction is allowed,
although not specified here.  Also, brackets are used for optional
components, and parentheses are used for grouping items.

\begin{figure}[h]
{\small
\begin{verbatim}
<TEMPORAL EXPRESSION> ::= [<RELATIVE>] <DATE PHRASE>
	              | [<RELATIVE>] <TIME PHRASE>
	              | [<RELATIVE>] <DATE PHRASE> ["at"] <TIME PHRASE>
	              | [<RELATIVE>] <TIME PHRASE> ["on"] <DATE PHRASE>
	              | "from" <DATE PHRASE> ("to"|"until") <DATE PHRASE> 
	              | "from" <TIME PHRASE> ("to"|"until") <TIME PHRASE> 

<RELATIVE> ::= "at" | "around" | "after" | "before" | "on" | "in"

<DATE PHRASE> ::= <DEICTIC> | <RELATIVE DATE> | <SPECIFIC DATE> | <ABSOLUTE DATE>

<DEICTIC> ::= "today" | "tomorrow" | "yesterday" | "now"

<RELATIVE DATE> ::= <SPECIFIER> <PERIOD>

<SPECIFIC DATE> ::= <DEMONSTRATIVE> <PERIOD>

<SPECIFIER> ::= "next" | "following" | "coming" | "the rest of" 
                | "the end of" | "this" | "this coming" | "all" 

<DEMONSTRATIVE> ::= "that" | "those" 

<PERIOD> ::= "day" | <WEEKDAY> | "week" | <MONTH> | "month" | "morning" 
             | "afternoon" | "lunchtime" | "evening" | "night"

<ABSOLUTE DATE> ::= [<WEEKDAY>] ["the"] <MONTH DATE> ["of" <MONTH>]
                    | [<WEEKDAY>] <MONTH> <MONTH DATE>
                    | ["in"] <MONTH>
                    | ["on"] <WEEKDAY>

<TIME PHRASE> ::= <HOUR_MIN> | <TIME OF DAY>

<HOUR_MIN> ::= <HOUR> <MINUTE>  [<TIME OF DAY>]

<TIME OF DAY> ::= "am" | "pm" | "in the morning" | "in the afternoon" 
                  | "lunch" | "night"
\end{verbatim}
}

\caption{BNF specification for system coverage of temporal expressions}
\label{BNF-coverage}
\end{figure}